\documentclass[10pt,twocolumn,letterpaper]{article}

\usepackage{iccv}
\usepackage{times}
\usepackage{epsfig}
\usepackage{graphicx}
\usepackage{amsmath}
\usepackage{amssymb}
\usepackage{subfigure}
\usepackage[accsupp]{axessibility}

\usepackage[pagebackref=true,breaklinks=true,colorlinks,bookmarks=false]{hyperref}

\iccvfinalcopy 

\def\iccvPaperID{****} 

\ificcvfinal\fi

\begin{document}

\title{Once Quantization-Aware Training: High Performance \\ Extremely Low-bit Architecture Search}

\author{{Mingzhu Shen$^1\ \ \ \ $ Feng Liang$^1\ \ \ \ $ Ruihao Gong$^1\ \ \ \ $ Yuhang Li$^1\ \ \ \ $ Chuming Li$^1\ \ \ \ $}\\
{Chen Lin$^2\ \ \ \ $ Fengwei Yu$^1\ \ \ \ $ Junjie Yan$^1\ \ \ \ $ Wanli Ouyang$^3\ \ \ \ $} \\
{$^1$Sensetime Research $\ \ \ \ ^2$University of Oxford $\ \ \ \ ^3$The University of Sydney} \\
{\tt\small shenmingzhu@sensetime.com, liangfeng@sensetime.com, wanli.ouyang@sydney.edu.au}
}

\maketitle

\begin{abstract}

Quantization Neural Networks (QNN) have attracted a lot of attention due to their high efficiency. To enhance the quantization accuracy, prior works mainly focus on designing advanced quantization algorithms but still fail to achieve satisfactory results under the extremely low-bit case. In this work, we take an architecture perspective to investigate the potential of high-performance QNN. 
Therefore, we propose to combine Network Architecture Search methods with quantization to enjoy the merits of the two sides. 
However, a naive combination inevitably faces unacceptable time consumption or unstable training problem. To alleviate these problems, we first propose the joint training of architecture and quantization with a shared step size to acquire a large number of quantized models. Then a bit-inheritance scheme is introduced to transfer the quantized models to the lower bit, which further reduces the time cost and meanwhile improves the quantization accuracy. Equipped with this overall framework, dubbed as Once Quantization-Aware Training~(OQAT), our searched model family, OQATNets, achieves a new state-of-the-art compared with various architectures under different bit-widths. 
In particular, OQAT-2bit-M achieves 61.6\% ImageNet Top-1 accuracy, outperforming 2-bit counterpart MobileNetV3 by a large margin of 9\% with 10\% less computation cost.
A series of quantization-friendly architectures are identified easily and extensive analysis can be made to summarize the interaction between quantization and neural architectures. Codes and models are released at \href{https://github.com/LaVieEnRoseSMZ/OQA}{https://github.com/LaVieEnRoseSMZ/OQA}

\end{abstract}

\section{Introduction}

Quantization Neural Networks (QNN) is a promising research direction to deploy deep neural networks on edge devices. Extensive efforts have been devoted to improving quantization performance with Quantization-Aware Training ~\cite{bulat2019xnor,zhou2016dorefa,gong2019differentiable,li2019additive,esser2019learned,bhalgat2020lsq+,hu2021opq} or Post-Training Quantization~\cite{nagel2020up,li2021brecq}. Recent studies~\cite{mishra2017wrpn,liu2020reactnet} investigate the quantization impact on architecture and thus take the architecture perspective to pursue high-performance quantized models with expert efforts and manual design. Compared with the laborious manual trials, the combination of Network Architecture Search (NAS) and quantization seems to be a more natural solution. 

\begin{figure} 
\subfigure[3-bit.]{
 \label{fig:sota_3bit_v2}     
\includegraphics[width=0.22\textwidth]{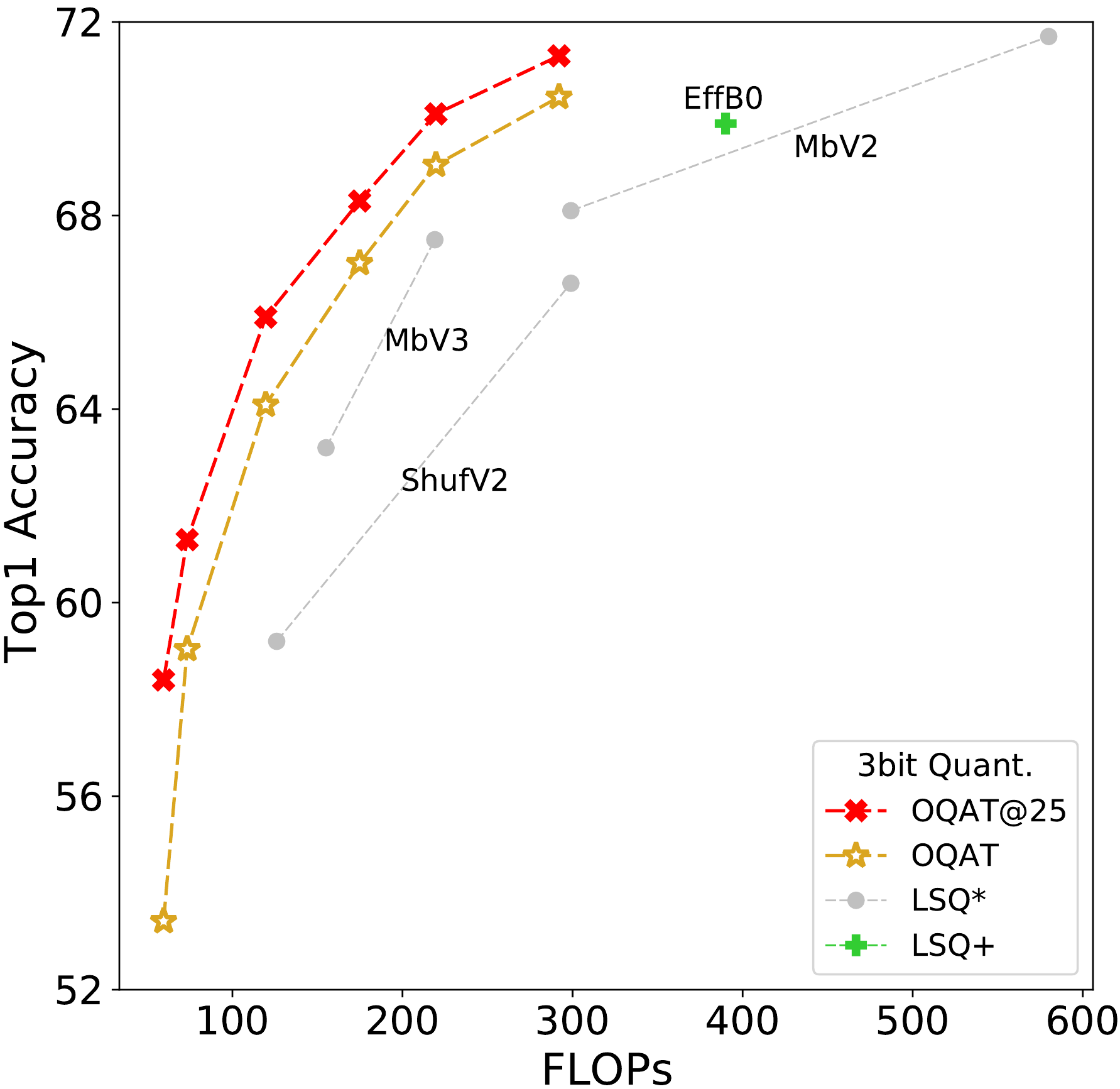}  
}     
\subfigure[2-bit.]{ 
\label{fig:sota_2bit_v2}     
\includegraphics[width=0.222\textwidth]{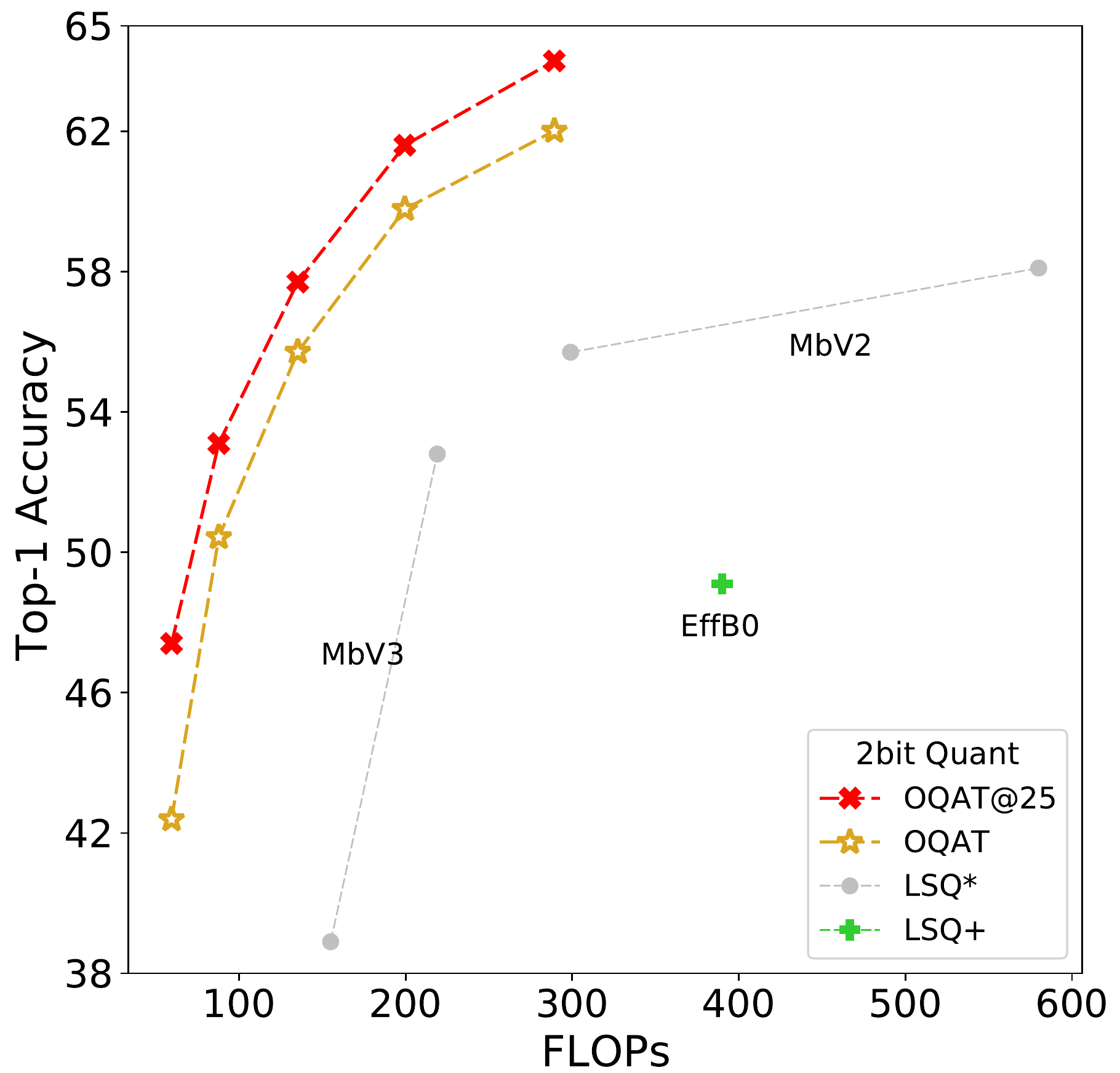}     
} 
\caption{Comparison with the state-of-the-art extremely low-bit neural networks on the ImageNet dataset. Our OQATNets achieve a new state-of-the-art under various  bit-widths.}
\label{fig:sota}     
\end{figure}

\begin{figure*}
    \centering
    \small
    \includegraphics[width=0.9\linewidth]{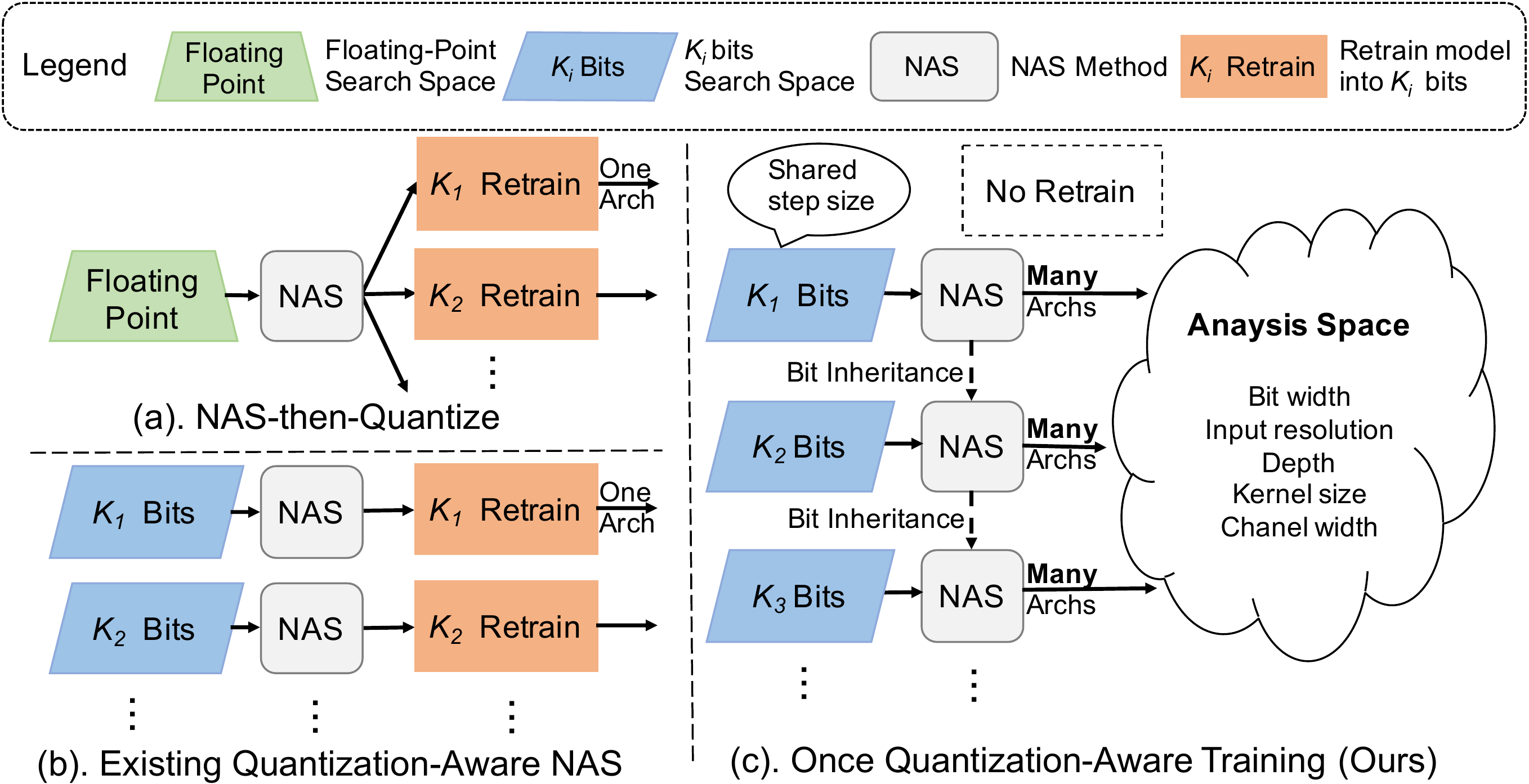}
    \caption{The overall framework of existing works on combining quantization and NAS methods. (a) NAS-then-Quanztize denotes directly converting the best searched floating-point architecture to quantization. (b) Existing Quantization-aware NAS first adopts a quantization-aware search algorithm to find a single architecture, then retrain the quantized weights and activation. (c) Our OQAT can search for many quantized compact models under various bit widths and deploy their quantized weights directly. }
    \label{overview}
\end{figure*}

The existing combination of NAS and quantization methods could either be classified as NAS-then-Quantize or Quantization-aware NAS as shown in Figure~\ref{overview}. NAS-then-Quantize~(Figure~\ref{overview}(a)) usually results in sub-optimal performance because the ranking order of full precision networks is not identical to that of quantized networks. Thus, this traditional routine may fail to get a good quantized model. Directly searching with quantized models' performance~(Figure~\ref{overview}(b)) seems to be an alternative. However, due to the instability brought by quantization-aware training, simply combining quantization and NAS results in inferior performance and sub-optimal quantized models as explained in~\cite{bulat2020bats}. Moreover, when quantized into 2-bit, the traditional training process is highly unstable and introduces very large accuracy degradation.

Furthermore, all the existing methods~\cite{wang2019haq, shen2019searching, bulat2020bats, guo2019single, wang2020apq} adopt a two-stage search-retrain scheme. Specifically, they first search for one architecture under the full-precision or low-bit setting and then retrain the model given the specific bit widths and architecture settings. This two-stage procedure undesirably increases the search and retrain cost if we have multiple deployment constraints and hardware bit widths.

To alleviate the aforementioned problems, we present Once Quantization-aware Training (OQAT), a framework that 1) trains quantized supernet with shared step size and deploys their quantized weights immediately without retraining, 2) progressively produces a series of quantized models under different bit-widths~(\eg 4/3/2 bit). Our approach leverages the recent NAS approaches which do not require retraining~\cite{yu2019universally, cai2019once, yu2020bignas, cheng2020scalenas} and combines it with quantization by shared step size. The search space for compact QNN includes kernel size, depth, width, and resolution. To provide a better initialization and transfer the knowledge of the higher bit-width to the lower bit-width, we propose a bit inheritance mechanism, which reduces the bit-width progressively to enable efficient searching for QNN under different quantization bit-widths. Benefiting from the non-retrain property and large search space under different bit widths, we conduct an extensive investigation on the interaction between neural architecture and model quantization, to shed light on the design principles of the quantized network.

Extensive experiments prove the effectiveness of our approach as shown in Figure~\ref{fig:sota}. Our searched quantized model family, OQATNets, achieves state-of-the-art results on the ImageNet dataset under 4/3/2 bit. In particular, our OQAT-2bit-M far exceeds the accuracy of 2-bit MobileNetV3@LSQ~\cite{esser2019learned} by a large 9\% margin while using 10\% less computation budget. Compared with the quantization-aware NAS method APQ~\cite{wang2020apq}, our OQAT-MBV2-4bit-L uses 43.7\% less computation cost while maintaining the same accuracy as APQ-B.

To summarize, the contributions of our paper are three-fold: 

\begin{itemize}

    \item Our OQAT is a quantization-aware NAS framework with the support of share step size to search for quantized compact models and deploy their quantized weights without retraining.
    
    \item We present the bit inheritance mechanism to reduce the bit-width progressively so that the higher bit-width models can guide the search and training of lower bit-width models.
    
    \item We provide insights into quantization-friendly architecture design. A comprehensive and systematic study reveals the favored quantization-friendly pattern under different bit widths.
    
\end{itemize}

\section{Related Work}


\paragraph{High-Performance QNN.} QNN has been widely used for efficiency in deployment. Extensive efforts have been devoted to improving the quantization performance of one given architecture. \cite{esser2019learned} proposes to learn step size in the quantization-aware training (QAT)~\cite{zhou2016dorefa,choi2018pact,bhalgat2020lsq+, li2021mqbench}, while some other research achieve this goal by reduce rounding error~\cite{nagel2020up,li2021brecq} with small computation cost in the post-training quantization (PTQ). These methods have achieved great progress in high bit widths like 8 with PTQ or under extremely low bit widths for heavy networks like ResNet18 with QAT. The quantization of efficient models~\cite{howard2019searching,li2020efficient} still causes large accuracy degradation. Recent works~\cite{mishra2017wrpn,liu2020reactnet} also find a strong correlation between architecture and quantization and intend to push the quantization performance by manual design.

\paragraph{Quantization-aware Network Architecture Search}
Inspired by recent progress of NAS studies like \cite{liu2018darts,guo2019single,liang2019computation,li2020improving,li2019lfs,zhou2020econas,yu2020bignas,cai2019once}, recent studies combine network quantization and NAS to automatically search for layer bit-width with given architecture or search for operations with given bit-width. HAQ~\cite{wang2019haq} focuses on searching for different bit widths for different layers in a given network structure and shows that some layers, which can be quantized to low bits, are more robust for quantization than others. AutoBNN~\cite{shen2019searching} utilizes the genetic algorithm to search for network channels and BMobi~\cite{phan2020binarizing} searches for the group number of different convolution layers under a certain 1-bit. SPOS~\cite{guo2019single} trains a quantized one-shot supernet to search for bit-width and network channels for heavy ResNet~\cite{he2016deep}. BATS~\cite{bulat2020bats} devises a binary search space and incorporates it within the DARTS framework~\cite{liu2018darts}. APQ~\cite{wang2020apq} trains a floating-point supernet and samples thousands of subnet for quantization to enable the transfer learning from the floating-point predictor to quantization predictor. Moreover, inherit a two-stage search-retrain scheme: once the best-quantized architectures have been identified, they need to be retrained for deployment. This procedure significantly increases the computational cost if we have different deployment constraints and hardware bit widths. 

\section{Preliminary}

\begin{table}
\centering
\caption{ImageNet performance under FP,4,3,2 bit with the same settings of quantization-aware training. Model-A and Model-B are two MobileNetV3-like models.}
\label{tab:reverse_resnet}
\begin{tabular}{l|lllll} 
\hline
Models      & FLOPs & FP   & 4-bit & 3-bit & 2-bit  \\
\hline
ResNet18    & 1813M & 71.0 & 71.0 & 70.2 & 67.6  \\
\hline
MobileNetV2 & 299M  & 72.0 & 71.3 & 68.2 & 55.7  \\
\hline
Model-A & 283M  & 75.0 & 72.3 & 68.9 & 54.6  \\
\hline
Model-B & 291M  & 75.5 & 72.3 & 67.8 & 49.5  \\
\hline
\end{tabular}
\end{table}

\subsection{Quantization}
Quantization maps the floating-point values into fix-point ones. In this paper, we choose uniform quantization since it is widely used in the practical deployment. Given a pre-defined $k$ bit, the weights and activation are quantized to corresponding signed $[-2^{k-1}, 2^{k-1}-1]$ and unsigned $[0, 2^{k}-1]$ integer range, respectively. The quantization function $Q$ can be formulated as: 
\begin{align}
\mathbf{v}^q = Q(\mathbf{v}, s) =\lfloor \mathrm{clip}( \frac{\mathbf v}{s},Q_{min},Q_{max}) \rceil \times s \label{quantize_forward},
\end{align}
where $\mathbf v$ represents the floating-point number and $\mathbf{v}^q$ is the quantized counterpart. $s$ is the step size for quantization mapping and $[Q_{min}, Q_{max}]$ represents the integer range.

To acquire the quantization accuracy, Post-Training Quantization and Quantization-Aware Training are two potential approaches. However, PTQ usually fails to achieve acceptable performance under the extremely low-bit setting~\cite{nagel2020up,li2021brecq}, which prevents us from revealing the internal quantization friendliness of a neural network. Thus we utilize QAT and adopt the Learned Step Size Quantization~\cite{esser2019learned} to ensure the reported accuracy represents the highest performance of a quantized network. LSQ directly optimizes the step size using the loss:
\begin{align}
& \frac{\partial \mathbf{v}^q}{\partial{\mathbf v}} \approx \mathbb{I}({\mathbf v}, Q_{min}\times s, Q_{max}\times s) \\
& \frac{\partial \mathbf{v}^q}{\partial{s}} \approx -\mathbb{I}(\frac{\mathbf{v}}{s}, Q_{min}, Q_{max}) + \lfloor \frac{\mathbf{v}}{s} \rceil,
\label{scale_grad}
\end{align}
where $\mathbb{I}(\mathbf v, Q_{min}\times s, Q_{max} \times s)$ means the gradient of $\mathbf{v}$ in the range of $(Q_{min}\times s, Q_{max} \times s)$ is approximated by 1, otherwise 0.

\subsection{Architecture Impact on Quantization}
As broadly investigated in the previous literature, architecture plays a crucial role in quantization performance. WRPN~\cite{mishra2017wrpn} improves the quantized accuracy by increasing the width of a neural network by manual design. ReActNet~\cite{liu2020reactnet} introduces a new activation module and brings new possibility for BNNs.

To make a deeper analysis into the interaction between architecture and quantization, we further conduct more validations. In Table~\ref{tab:reverse_resnet}, we compare two widely used models ResNet18 and MobileNetV2. Although MobileNetV2 surpasses the accuracy of ResNet18 in full-precision, the accuracy of 2-bit MobileNetV2 reduces significantly. It is common sense that compact models with depthwise convolution like MobileNetV2 tend to be more sensitive to quantization than heavy networks like ResNet18. We further present two MobileNetV3-like models with different depth, width, and kernel size settings. With a similar computation budget, Model-B surpasses Model-A with 0.5\% accuracy gain in floating-point settings, while the accuracy of Model-B is 5\% lower than Model-A in 2-bit. It indicates that different architecture results in different quantization performances even when they share similar FP accuracy. 

\section{Methodology}

In this section, we present Once Quantization-Aware Training (OQAT), a framework that jointly trains quantization and search architectures within a huge search space. 

\subsection{Once Quantization-Aware Training}

We first review two prevalent NAS + Quantization routes: (1) \textit{NAS-then-Quantize} searches full precision model performance and then quantizes the optimal FP model as shown in Figure~\ref{overview}(a). This type of method may result in sub-optimal architecture since the quantization-friendliness is not considered during the search process. (2) \textit{Quantization-aware NAS} proposes to incorporate quantization-aware performance in the NAS process as shown in Figure~\ref{overview}(b). However, the simple combination of NAS and quantization results in unstable training or inferior performance as explained in BATS~\cite{bulat2020bats}. To stabilize the training process, \cite{bulat2020bats} proposes to only quantize activation which does not search the global quantized architectures. 
Moreover, both NAS-then-Quantize and Quantization-aware NAS require a two-stage search-retrain scheme, which is unaffordable if we have multiple deployment requirements.

To this end, we propose Once Quantization-Aware Training~(OQAT), a framework that jointly trains quantization and searches architecture with a large search space inspired by recent non-retrain NAS methods~\cite{yu2018slimmable,yu2019universally,cai2019once,yu2020bignas}. Specifically, a quantized supernet with the largest possible depth (number of blocks), width (number of channels), kernel size, and input resolution is trained. Then a subnet is obtained from parts of the supernet with depth, width, and kernel size smaller than the supernet. The subnet uses the well-trained parameters of the supernet with simple BatchNorm calibration~\cite{yu2019universally} for direct deployment without further retraining. 

The overall procedure of OQAT is illustrated as follows: \underline{Step 1}, quantized supernet training (Section~\ref{Sec:framework}): train a $k$-bit supernet by learning the weight parameters and step size simultaneously. \underline{Step 2}: given a constraint on computational complexity, search for the architecture with the highest quantization performance on the validation dataset. If $k=n$, the whole process is finished. \underline{Step 3}, bit inheritance (Section~\ref{sec:bit-inheritance}): Use the weight and quantization parameters of the $k$ bit supernet to initialize the weight and quantization parameters of the $k-1$ bit supernet. \underline{Step 4}: $k\leftarrow k-1$ and Go to step 1. 




Joint training of quantization and architecture does not come as free lunch as it might appear, but is more subtle and involves new designed techniques. Unlike the floating-point supernet training~\cite{cai2019once, yu2020bignas}, the weights and activation values are quantized with Eq.~\ref{quantize_forward} for the quantized supernet training. As explained in LSQ~\cite{esser2019learned} and recent quantization robustness papers~\cite{shkolnik2020robust,alizadeh2020gradient}, learned step size is sensitive to training, and optimal step size is critical for final quantization accuracy. The design of the learned step size should be carefully handled. Under ultra-low bit-width, the traditional QAT becomes highly unstable and introduces large accuracy degradation. How to produce low-bit quantized supernet with high performance efficiently is also crucial. 

\subsection{Shared Step Size}
\label{Sec:framework}

\begin{figure} 
\subfigure{
 \label{fig:activation_point_linear}     
\includegraphics[width=0.225\textwidth]{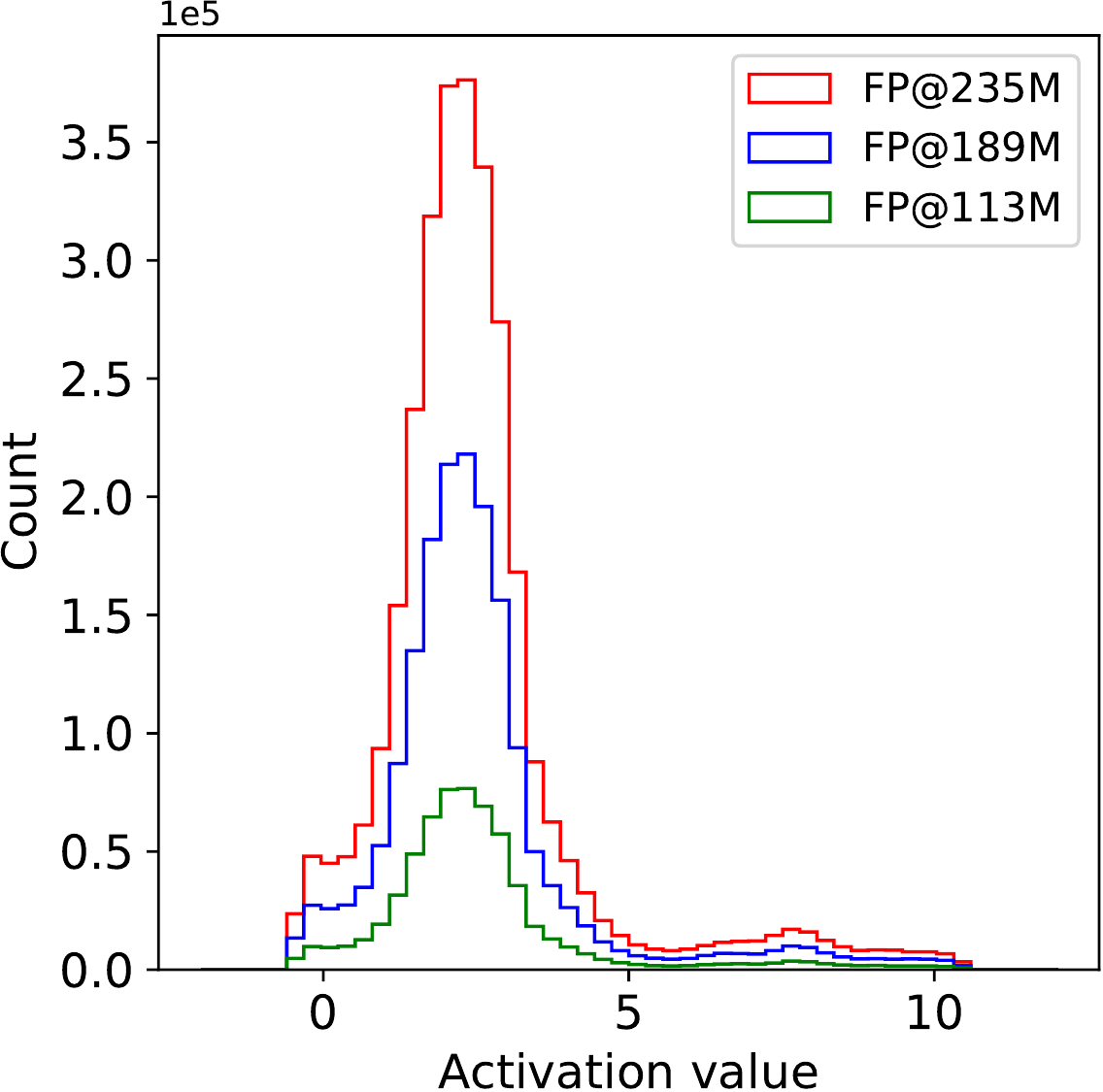}  
}     
\subfigure{ 
\label{fig:weight_point_linear}     
\includegraphics[width=0.215\textwidth]{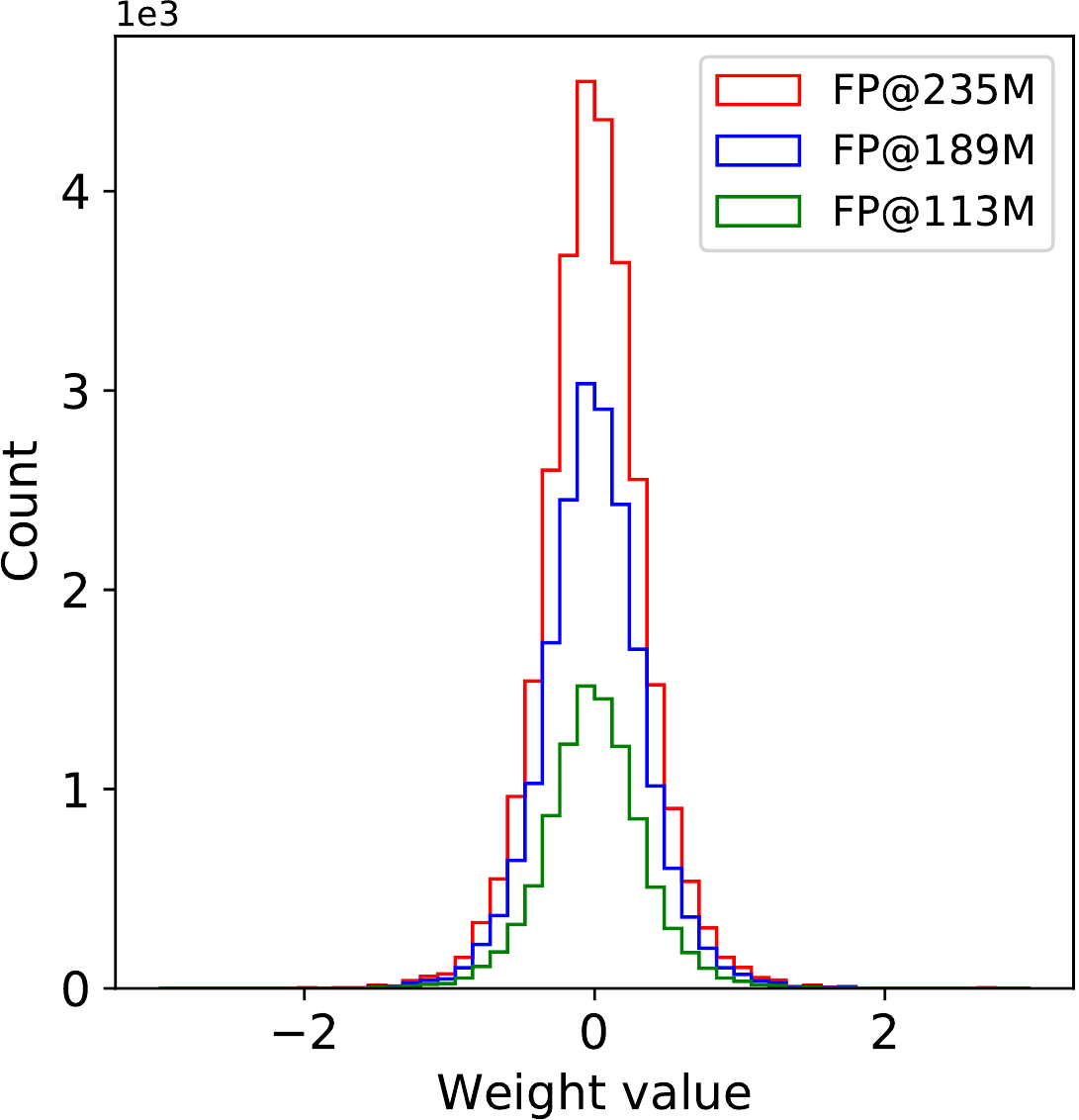}     
} 
\caption{The rationality of shared step size. We randomly pick three subnets from the floating-point (FP) supernet. The subnet is denoted as FP@FLOPs. We depict the activation and weights of the second pointwise convolution layer in the third stage. We find that although the subnets have quite different FLOPs, their activation and weights share almost the same range (x-axis).}
\label{fig:shared_step_size}     
\end{figure}

Quantization step size is essential for quantization error, as it balances the rounding and clipping error. 
There are 3 potential approaches to handle learned step size:
\begin{enumerate}
    \item Assign every subnet with unique step sizes for both weights and activation. However, it is impossible to store the huge amount of step sizes for $10^{20}$ subnets.
    \item Apply switchable step size for every possible candidate choice in one layer inspired by Switchable BN~\cite{yu2018slimmable} and Switchable Clipping Level~\cite{jin2019adabits}. For example, if the kernel size of one layer has three search settings with $\{3, 5, 7\}$, the number of step sizes is also set to three.
    \item Equip every layer with only one step size for weights and one step size for activation.
\end{enumerate}

We first plot the distribution of weights and activation of floating-point supernet and find that the range and distribution are similar among different subnets as shown in Figure~\ref{fig:shared_step_size}. It indicates that shared step size is perhaps enough to share across subnets. We also experiment with the latter two methods to support quantized supernet training with elastic kernel size and they indeed result in similar results (see Appendix). And for the simplicity of efficient training, we propose to share the step size across one layer.

Therefore, the forward pass of the quantized supernet training is defined as follows. Given a floating-point convolution layer for candidate architecture $arch_i$, the floating-point weights $\mathbf w$ and activation $\mathbf{a}$ and the learned step size $s_{\mathbf a}$ and $s_{\mathbf w}$ are shared through all candidate architectures in this layer. In the forward pass, the weights $\mathbf{w}_{i}$ and activation $\mathbf{a}_{i}$ are cropped respectively. The learned quantization function defined in Eq.~\ref{quantize_forward} is used to quantize weights and activation into $\mathbf{w}_i^q$ and $\mathbf{a}_i^q$. The forward pass for a quantized convolution layer is then given by $\mathbf{y}=\mathbf{w}_i^q*\mathbf{a}_i^q$ where $*$ is the convolution.

\subsection{Bit Inheritance}
\label{sec:bit-inheritance}

\begin{table}
\caption{
The accuracy of the biggest model with adaptation to lower bit-width. 4-bit denoted the 4-bit models after individual QAT training, OQAT-4bit denotes quantized supernet at 4-bit. Calib denotes BatchNorm and step size calibration with little computation overhead. 4/3 represents the bit width of weights and activation is 4 and 3 respectively.
}
\centering
\small
\begin{tabular}{l|c|cccc}
\hline
Models & Methods & 4/4 & 4/3 & 3/4 & 3/3\\
\hline
4-bit & Calib & 75.0\% & 72.9\% & 69.2\% & 65.3\% \\
\hline
OQAT-4bit & Calib & 75.5\% & 73.8\% & 72.1\% & 69.1\% \\
\hline
\end{tabular}
\label{tab:robustness}
\end{table} 

When the bit-width is lower than 3, the traditional quantization-aware training (QAT)~\cite{kim2019qkd, bhalgat2020lsq+} process is highly unstable and introduces large accuracy degradation for the 2-bit model. Besides, despite the higher efficiency of OQAT compared with existing methods, the time cost is still unacceptable when we need to train a supernet for each bit-width.

To compensate for the disadvantages, we devise a bit inheritance procedure, which serves as a good initialization for low-bit models and improves the final QAT accuracy efficiently. We propose that in the initialization for $k-1$ bit supernet, the weight parameters and step size are inherited and calibrated from $k$ bit supernet. Specifically, the step size of $k-1$ bit-width is calibrated with double the original step size. With this simple principle, we can promise a bounded error when transferring the bit-width between supernets. Given a convolution layer in the $k$ bit quantized model, we denote $s_k$ as the step size of this layer, and $s_{k-1} = 2s_k $ as the doubled step for the $k-1$ bit model. We use $\mathbf{w}$ and $N_\mathbf{w}$ to denote the weights of this layer which is inherited from $k$ to $k-1$. Next, we show that the $L_1$ distance of $Q(\mathbf{w}, s_k)$ and $Q(\mathbf{w}, s_{k-1})$ is bounded by $N_\mathbf{w} \cdot s_k$. It means the initialized $Q(\mathbf{w}, s_{k-1})$ has a bounded distance with the well-trained $Q(\mathbf{w}, s_k)$. For each $w_i$, we have:

\begin{equation}
||Q(\mathbf{w}, s_k)-Q(\mathbf{w}, s_{k-1})||_1 \leq N_w \cdot |s_k|.
\end{equation}

The detailed proof can be seen in the Appendix. The theorem indicates that our bit inheritance scheme can provide a better initialization for the later finetuning.

\paragraph{Effectiveness of Bit Inheritance}
To validate the effectiveness of bit inheritance, we first directly use it to convert the 4-bit quantized supernet to 3 bit with simple BN and step size calibration. We surprisingly find that the accuracy has already outperformed the result with a complete QAT for the same network (see Table~\ref{tab:robustness}). Besides the bounded error brought by the bit inheritance itself, we conjecture that our shared step size and shared parameter training also contribute to the robust transfer. Further, we finetune the inherited 3-bit supernet for just one epoch, the accuracy can effortlessly improve by 2.6\%, which also shows that the inherited supernet has already been around the global minimum. Even for the lower bit-width such as 2-bit, it can achieve accuracy close to 3-bit with just a few more epochs' finetuning after inheritance (see Table~\ref{tab:bitshrinking}). All the evidence proves that our bit inheritance technique is efficient and effective for producing high-performance extremely low-bit neural networks.

\begin{table}
\caption{
The accuracy of the biggest model in (QAT) and progressive bit inheritance. Start and End denote the accuracy at the first epoch and the end of the training.
}
\centering
\small
\begin{tabular}{l|ccc}
\hline
Methods & 4/4 & 3/3 & 2/2\\
\hline
QAT@Start & 48.1\% &  23.2\% & 0.8\% \\
\hline
QAT@End & 75.1\% & 72.1\% & 56\% \\
\hline
Bit-Inheritance@Start & - & 71.7\% & 49.3\% \\
\hline
Bit-Inheritance@End & - & 72.7\% & 64.5\% \\
\hline
\end{tabular}
\label{tab:bitshrinking}
\end{table}


\begin{table*}
\centering
\small
\caption{The comparison of the search cost and retrain cost with existing quantization-aware NAS methods. $N$ denotes the number of models to be deployed. The total cost is calculated with $N=40$.}
\label{tab:timecost}
\begin{tabular}{lccccc} 
\hline
Methods & SPOS~\cite{guo2019single}  & BMobi~\cite{phan2020binarizing} & BATS~\cite{bulat2020bats} & APQ~\cite{wang2020apq} & \textbf{OQAT} \\
\hline
search cost (GPU hours) & $288+24N$ & $29N$ &  $6N$  & $2400+0.5N$ & \textbf{1200+0.5N}  \\
retrain cost (GPU hours) & $240N$  & $256N$ & $75N$ & $30N$       & \textbf{0}   \\
\hline
total cost (GPU hours) & $10.8k$ & $11.4k$ & $3.2k$ & $3.6k$ & \textbf{1.2k} \\
\hline
\end{tabular}
\end{table*}

\begin{table*}
\centering
\small
\caption{The best architecture for full-precision~(FP) is not the best for quantization and bit inheritance can effectively improve quantization accuracy even compared with long-time QAT. The two architectures share similar FLOPS and FP accuracies (columns 2, 3) but show different quantization friendliness. And even we finetune the model with a complete QAT for 150/500 epochs (column 5, 6), the accuracy is still far from that of the subnet directly sampled from BI 2-bit supernet (column 4).} 
\label{tab:ablation}
\begin{tabular}{cccccc} 
\hline
Architecture            & FLOPs & \begin{tabular}[c]{@{}c@{}}Top-1 Acc.(\%)\\ In FP Supernet\end{tabular} & \begin{tabular}[c]{@{}c@{}}Top-1 Acc.(\%)\\In BI 2-bit Supernet\end{tabular} & \begin{tabular}[c]{@{}c@{}}Top-1 Acc.(\%)\\QAT@150\end{tabular} & \begin{tabular}[c]{@{}c@{}}Top-1 Acc.(\%)\\QAT@500\end{tabular}  \\ 
\hline
Pareto model from FP supernet & 144   & 73.6\% & 54.4\% & 28.1\% & 43.5\% \\ 
Pareto model from 2-bit supernet      & 142   & 73.4\%  & \textbf{56.1}\%  & 33.2\% & 47.2\%  \\
\hline
\end{tabular}
\end{table*}

\section{Experimental Analysis}

\subsection{Experimental settings}
\paragraph{Implementation details.} We evaluate our method on the ImageNet dataset~\cite{deng2009imagenet}. If not specified, we follow the standard practice for quantized models~\cite{kim2019qkd, gong2019differentiable} on quantizing the weights and activation for all convolution layers except the first convolution layer, last linear layer, and the convolution layers in the SE modules~\cite{hu2018squeeze}. 
To fairly compare with quantized compact models, the FLOPs and BitOPs are defined as follows. Denote the FLOPs of the FP layer by $a$, the BitOPs of $m$-bit weight and $n$-bit activation quantized layer is $mn\times a$ following~\cite{zhou2016dorefa,li2019additive,phan2020binarizing,bulat2020bats,wu2018mixed}.  Further details can be found in Appendix.

\paragraph{Search space.} Our search space is based on MobileNetV2 (MBV2)~\cite{sandler2018mobilenetv2} and MobileNetV3 (MBV3)~\cite{howard2019searching}, which has the flexible input resolution, filter kernel size, depth (number of blocks in each stage), and width (number of channels). Our search space consists of multiple stages. Each stage stacks several inverted residual blocks. Further details about search space can be found in the Appendix. Unless otherwise noted, all results are sampled from the MBV3 search space denoted as OQAT, OQAT-MBV2 represents the MBV2 search space. Further details can be found in Appendix.

\paragraph{Architecture search of quantized supernet.} For quantization-friendly analysis, we directly evaluate the sampled subnets from the supernet without further retraining. We randomly sample 10K candidate architectures from the supernet with the FLOPs of the corresponding floating-point models ranging from 50M to 300M (2K in every 50M interval). The no-retraining property enables us to conduct the analysis within a huge search space. For one specific model deployment requirement, we exploit a coarse-to-fine architecture selection procedure, similar to~\cite{yu2020bignas}. We first randomly sample 500 candidate architectures from the supernet within the FLOPs range of $\pm 10\%$. After obtaining the good skeletons (input resolution, depth, width) in the pareto front, we randomly perturb the kernel sizes to further search for better architectures.

\subsection{The efficiency of OQAT}
 In Table~\ref{tab:timecost}, we compare the search cost and the retrain cost of OQAT with existing methods. The search cost is defined as the time cost of the supernet training and the search process to get the final $N$ searched models under latency targets. The retrain cost is defined as the training cost to get the final accuracy of the searched architecture. When $N$ is larger than 5, the retrain cost of SPOS~\cite{guo2019single} and BMobi~\cite{phan2020binarizing} will surpass the total cost of OQAT. Compared with APQ~\cite{wang2020apq}, our method can reduce half of the total cost. APQ needs to train an FP supernet and sample thousands of FP subnets to perform quantization-aware training, and thus requires the transfer learning from the floating-point predictor to the quantization predictor. Our OQAT only needs to train one quantized supernet and support a huge search space with over $10^{20}$ subnets that can be directly sampled from supernet without retraining. Thus, the average computational cost is relatively low.

\subsection{Ablation study}
In Table~\ref{tab:ablation}, we validate the advantages of joint training and bit inheritance. The model in the first row is selected from the pareto front with the floating-point accuracy which corresponds to NAS-then-Quantize, and the latter is with the 2-bit accuracy as OQAT. With similar FLOPs, similar floating-point accuracy, OQAT surpasses the accuracy of Nas-then-Quantize with $1.7\%$ in 2-bit accuracy. We also perform QAT with 150 epochs and the joint training results in over $5\%$ accuracy improvement, which verifies the effectiveness of OQAT in finding quantization-friendly architectures. Although 2-bit models benefit from QAT with more epochs (e.g., 500 epochs), the achieved accuracy is still far from that directly sampled from the bit inheritance (BI) supernet. The huge accuracy improvement verifies that bit inheritance is a better practice compared with the existing routine of quantization-aware training because it alleviates the problem that quantized compact models with low bit-width are highly unstable to train.

\subsection{Comparison with existing architectures}
Benefiting from joint quantization and NAS with a large search space, as well as the bit inheritance for low-bit quantized supernet, we get high-performance OQATNets under extremely low bit-width. As shown in Figure~\ref{fig:sota_3bit_v2} and Figure~\ref{fig:sota_2bit_v2}, OQATNets can be directly deployed for its superior accuracy, while the accuracy can be further improved by finetuning with 25 epochs denoted by OQAT@25.

We implement LSQ~\cite{esser2019learned} denoted by LSQ* to construct strong baseline and compare with another strong quantization methods LSQ+~\cite{bhalgat2020lsq+}. Our OQATNets outperforms multiple architectures like MobileNetV2~\cite{sandler2018mobilenetv2}, EfficientNet-B0~\cite{tan2019efficientnet} and MobileNetV3~\cite{howard2019searching} under all bit-widths we implements. 
\textbf{3 bit:} Our OQAT-3bit-L can also outperforms the accuracy of EfficientB0 by 1.3\% with 15\% FLOPs.
\textbf{2 bit:} Our OQAT-2bit-M requires less FLOPs but achieves significantly higher Top-1 accuracy (61.7\%) when compared with MobileNetV3@LSQ* (52.8\%) and MobileNetV2@LSQ* (55.7\%). The results verify that the OQAT results in quantization-friendly compact models.

\begin{figure}[ttb] 
\centering   
\subfigure[NAS-then-Quantize.] {
 \label{fig:FP_pareto}     
\includegraphics[width=0.22\textwidth]{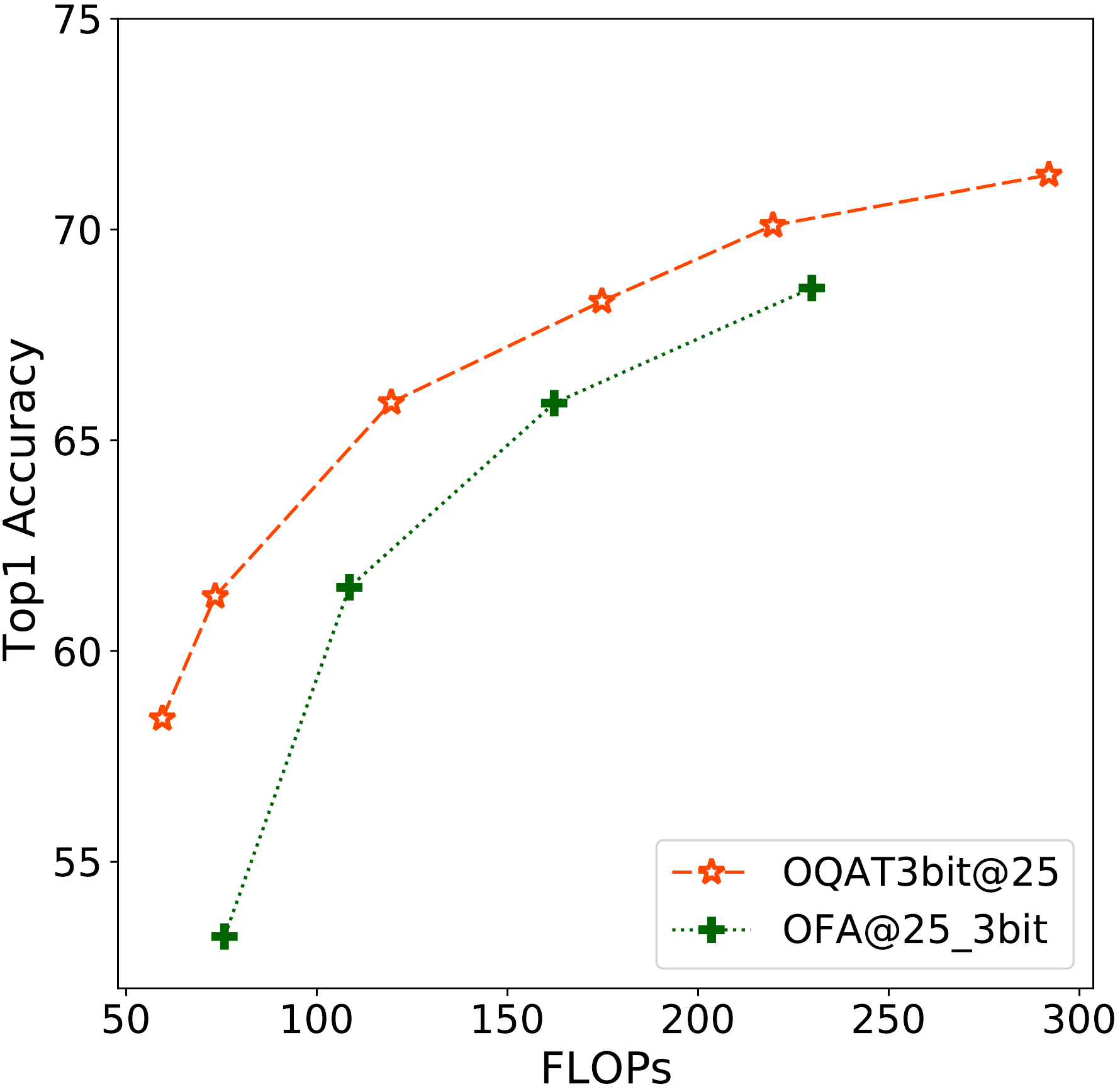}  
}     
\subfigure[Pareto models.] { 
\label{fig:3bit_pareto}     
\includegraphics[width=0.225\textwidth]{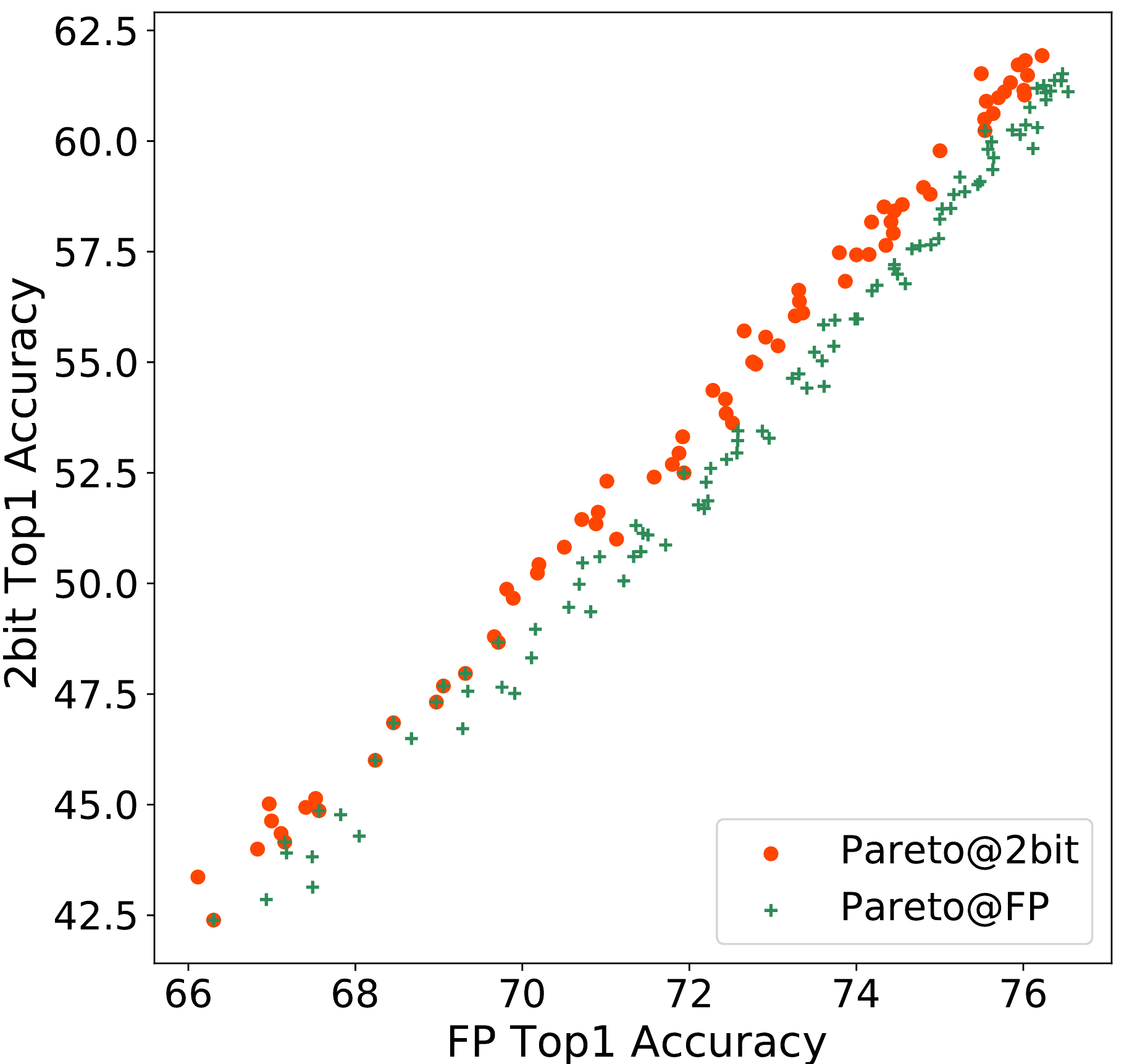}     
}    
\caption{Comparison of the parato models of NAS-then-Quantize and OQAT. The FLOPs of the corresponding floating-point architectures are used. The accuracy of Subnets@FP/3bit/2bit and Pareto@FP/3bit/2bit is obtained in the correponding FP/3bit/2bit supernet.}
\label{fig:nas-then-quantize}     
\end{figure}

\paragraph{NAS-then-Quantize.} We further compare with searched model produced by NAS-then-Quantize procedure (Figure~\ref{overview}(a)). As shown in Figure~\ref{fig:nas-then-quantize}(a), we quantize the pareto models of the OFA\cite{cai2019once} floating-point supernet and compare it with the pareto models of our OQATNets. In the comparison under 3 bit, the pareto curve of our OQAT is far above that of the NAS-then-Quantize. 

In Figure~\ref{fig:nas-then-quantize}(b), we sample 10k subnets from the search space, and we validate these architectures from the FP supernet and 2-bit supernet. The pareto front of the subnets denoted as Parato@FP are selected with the FP accuracy and Pareto@2bit are selected with the 2-bit accuracy. Pareto models as those models on the pareto front of the cost/accuracy trade-off curve. With the same accuracy of the floating-point models, the accuracy of the model from the 2-bit pareto is higher than the model from the FP pareto. If our target is to search architectures for the quantized models, searching from the quantized supernet as our OQAT did is better than searching from FP supernet and then quantized. 

\paragraph{Existing Quantization-aware NAS.} In Table~\ref{tab:quantization-aware-nas}, we compare our OQATNets model family with several searched models from existing quantization-aware NAS methods, named HAQ~\cite{wang2019haq}, SPOS~\cite{guo2019single}, BMobi~\cite{phan2020binarizing}, BATS~\cite{bulat2020bats} and APQ~\cite{wang2020apq}. 

Our OQATNets model family shows great advantages over existing weight-sharing methods corresponding to the paradigm of Figure~\ref{overview}(b). While SPOS~\cite{guo2019single} focuses on the search of network channels and bit-width of heavy ResNet~\cite{he2016deep}, we focus on the search of compact models with fixed bit-width and achieve better results with fewer FLOPs. BMobi~\cite{phan2020binarizing} and BATS~\cite{bulat2020bats} did not provide the results for 2-bit, 3-bit or 4-bit. Therefore, we would like not to directly compare our approach with BMobi and BATS, because the results are obtained from different bit-widths. However, if only the FLOPs-accuracy trade-off is concerned, our OQAT with a higher bit-width can be a better solution. APQ utilizes transfer learning from FP predictor to quantized predictor which may bring proxy problems. Our OQAT-MBV2-4bit-L uses 43.7\% less computation cost while maintaining the same accuracy as APQ-B.

\begin{table}[t]
\centering
\caption{Quantization-aware NAS performance under different bit-widths on ImageNet dataset. \emph{Bit (W/A)} denotes the bit-width for both weights and activation. The number of bit for different layers is different for 
SPOS~\cite{guo2019single} and APQ~\cite{wang2020apq}. BMobi~\cite{phan2020binarizing}, BATS~\cite{bulat2020bats}, 
and OQAT use the same bit-width for different layers. L and M are short for for Large and medium model size.}
\label{tab:quantization-aware-nas}
{\small
\begin{tabular}{lccc} 
\hline
Models            & Bit(W / A)  & BitOPs(G) & Top-1   \\ 
\hline
SPOS-ResNet34    & \{1, 2, 3, 4\}            & 13.11   & 71.5            \\
SPOS-ResNet18    & \{1, 2, 3, 4\}         & 6.21    & 66.4\%            \\ 
\hline
BATS-$2\times$         & 1          &  9.92    & 66.1\%            \\
BATS-$1\times$         & 1       & 6.30    & 60.4 \%           \\
\hline
BMobi-M1        & 1     & 3.97         & 59.3\%            \\
BMobi-M2     & 1         & 2.11        & 51.1\%            \\
\hline
\textbf{OQAT-3bit-L}  & 3     & 3.07  & \textbf{71.3\%}   \\
\textbf{OQAT-3bit-M}    & 3    & 1.92  & \textbf{68.3\%}   \\
\textbf{OQAT-2bit-M}     & 2     & 1.21  & \textbf{61.7\%}   \\
\hline
APQ-B          & \{4, 6, 8\} &  16.5    & 74.1\%           \\
APQ-A          & \{4, 6, 8\}  & 13.2    & 72.1\%            \\
 
\hline
\textbf{OQAT-MBV2-4bit-L}     & 4    & 9.28  & \textbf{74.1\%}   \\
\textbf{OQAT-MBV2-4bit-M}      & 4      & 6.85   & \textbf{72.4\%}   \\
\hline
\end{tabular}
}
\end{table}

\begin{figure*} 
\subfigure[Different bit-widths.] {
 \label{fig:bit-width_depth}     
\includegraphics[width=0.23\textwidth]{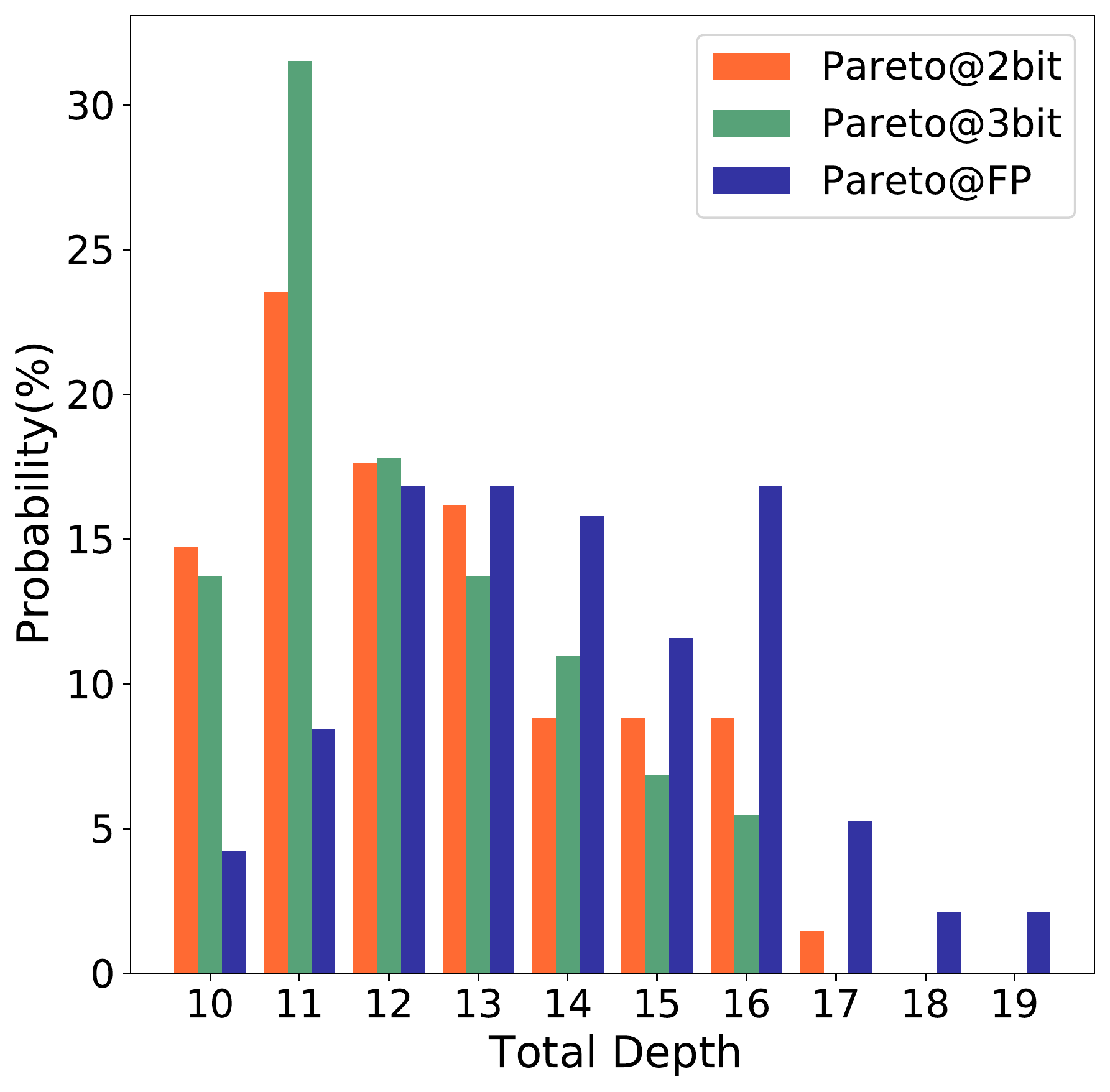}  
}    
\subfigure[Different bit-widths.] {
 \label{fig:bit-width_width}     
\includegraphics[width=0.23\linewidth]{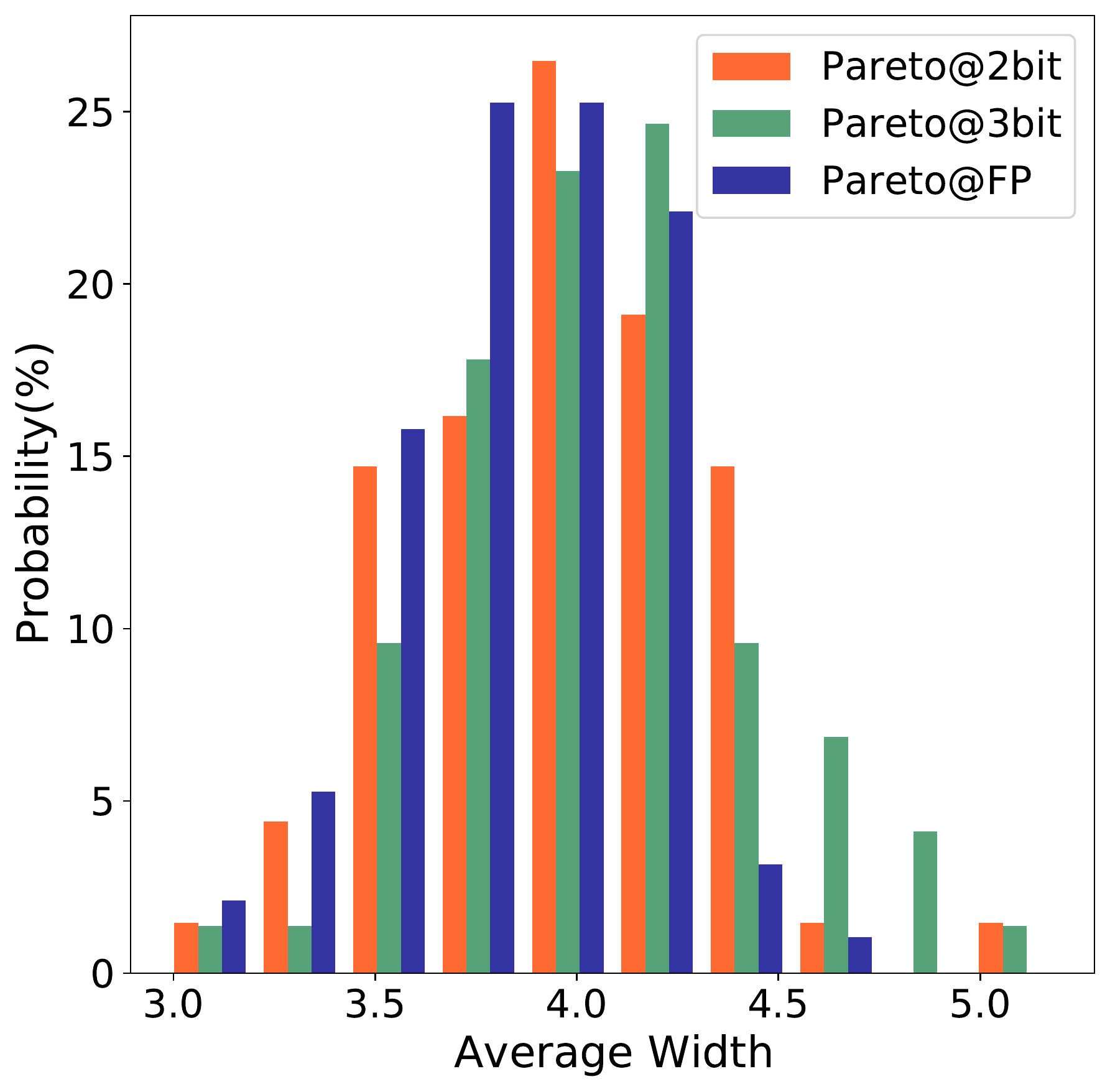}  
}
\centering    
\subfigure[FLOPs.] { 
\label{fig:flops}     
\includegraphics[width=0.233\linewidth]{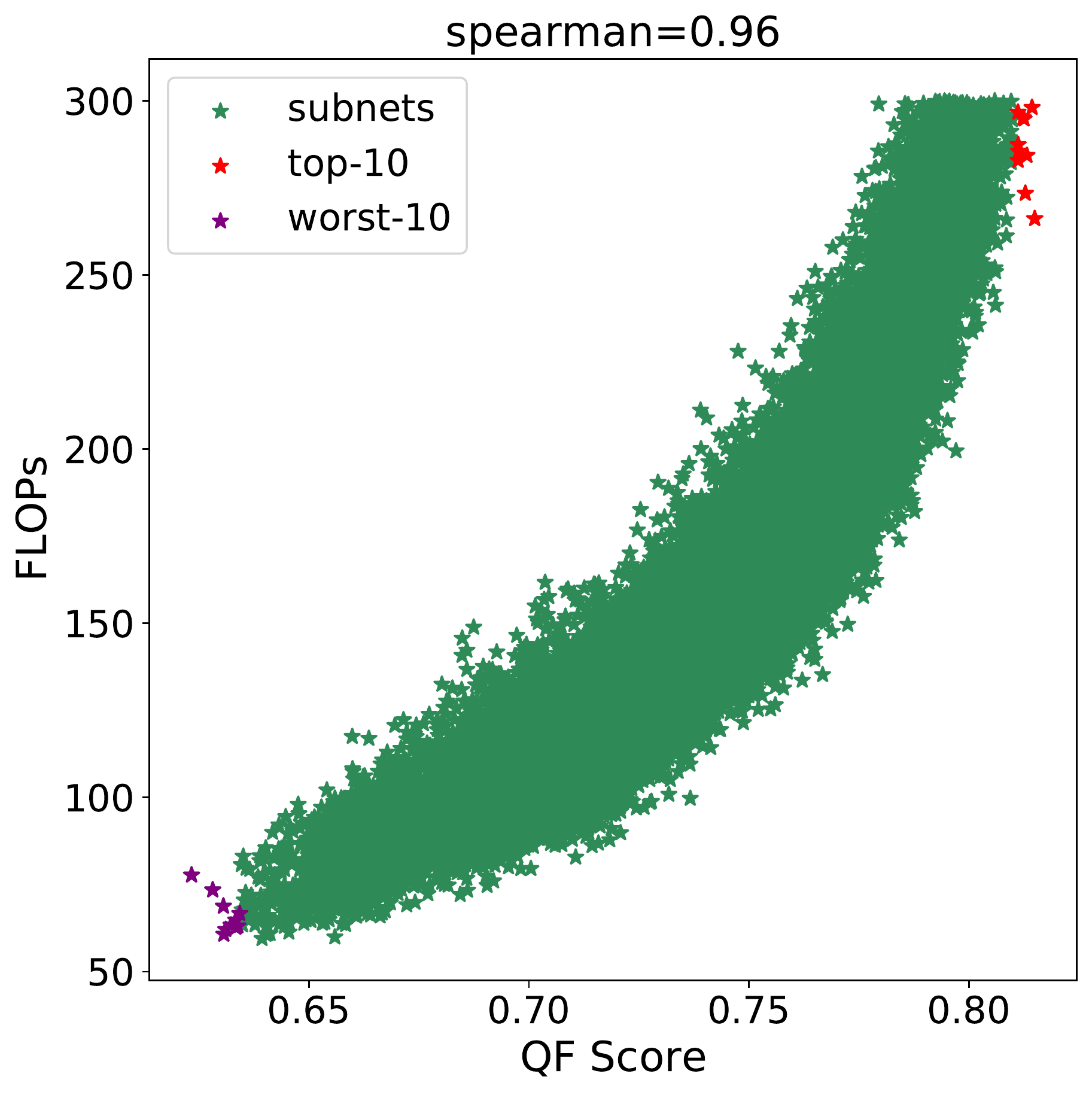}     
} 
\centering    
\subfigure[Resolution.] { 
\label{fig:resolution}     
\includegraphics[width=0.23\linewidth]{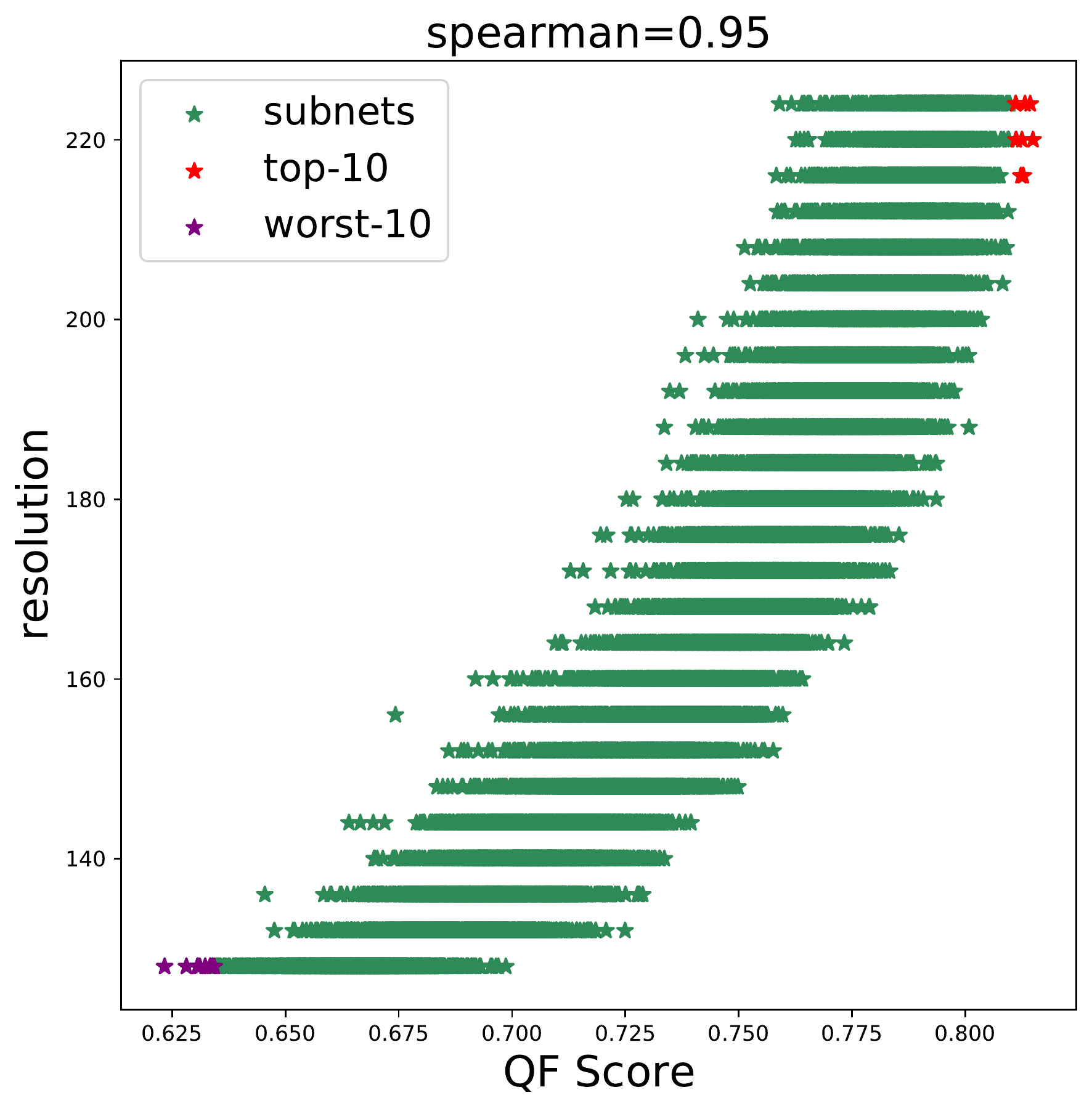}     
} 
\subfigure[Given computation budget.] { 
\label{fig:computation_budget_depth}     
\includegraphics[width=0.23\linewidth]{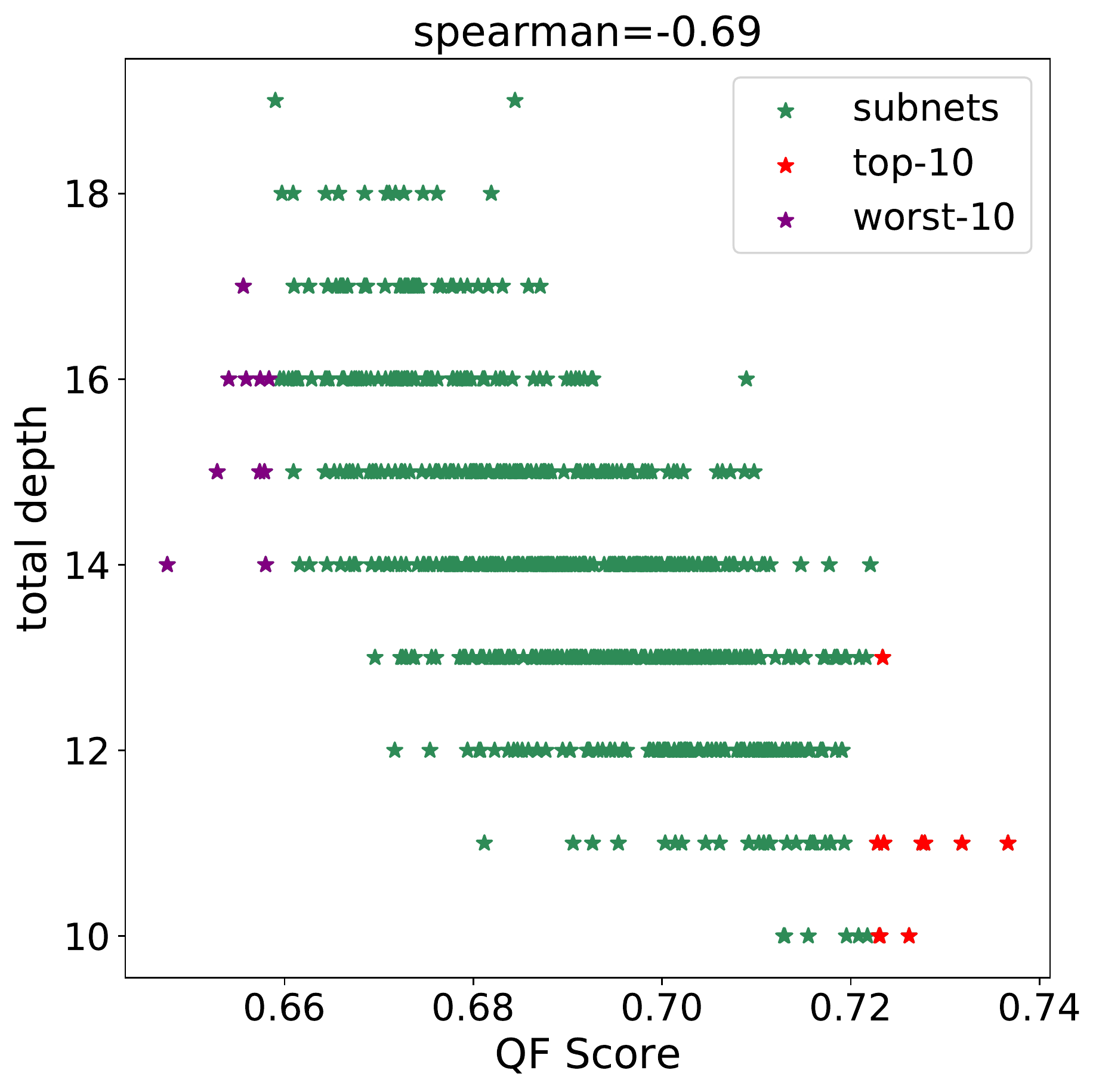}     
}    
\subfigure[Given computation budget.] { 
\label{fig:computation_budget_resolution}     
\includegraphics[width=0.235\textwidth]{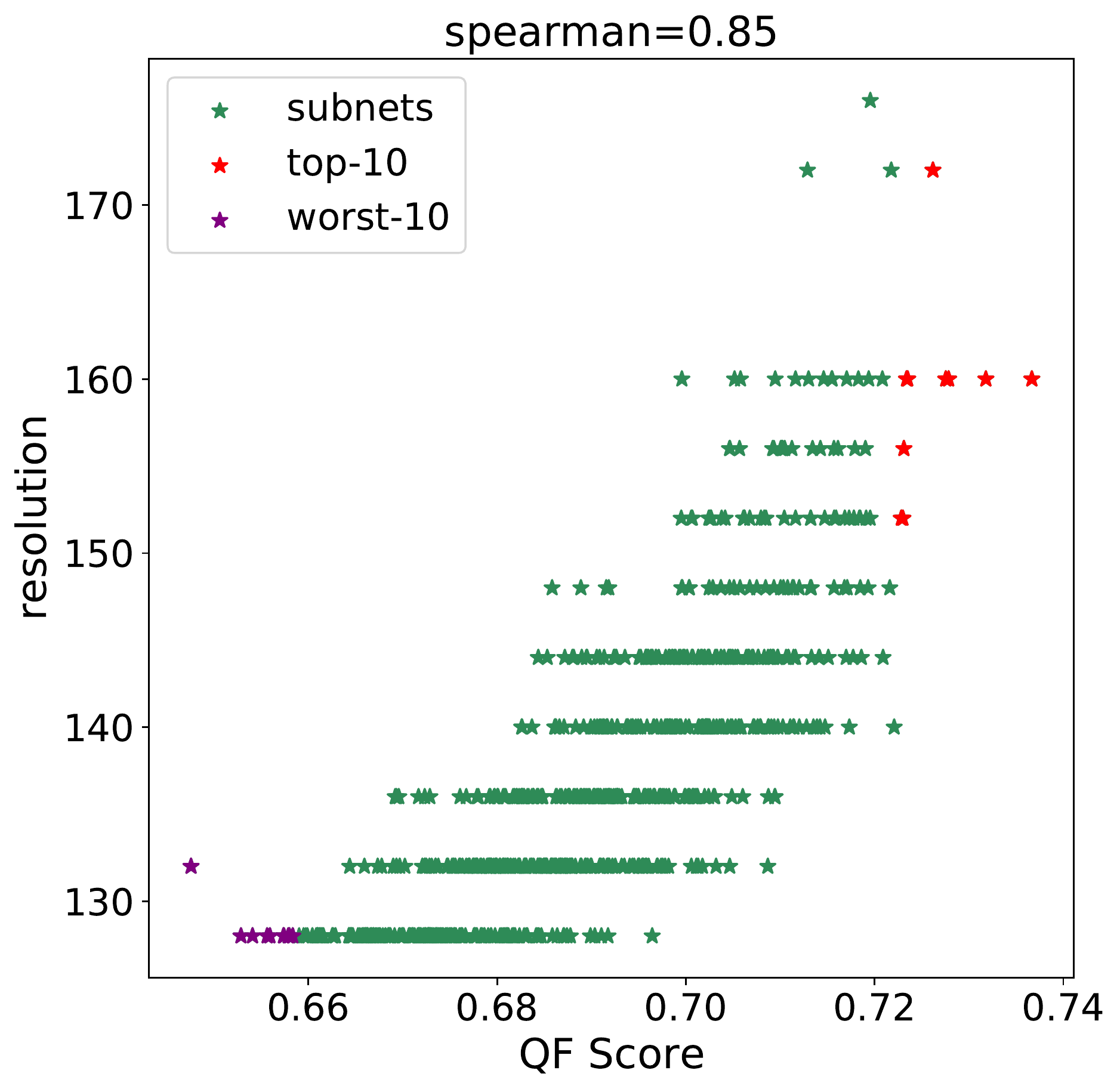}     
}  
\subfigure[Given floating-point accuracy.] { 
\label{fig:fp_acc_width}     
\includegraphics[width=0.23\linewidth]{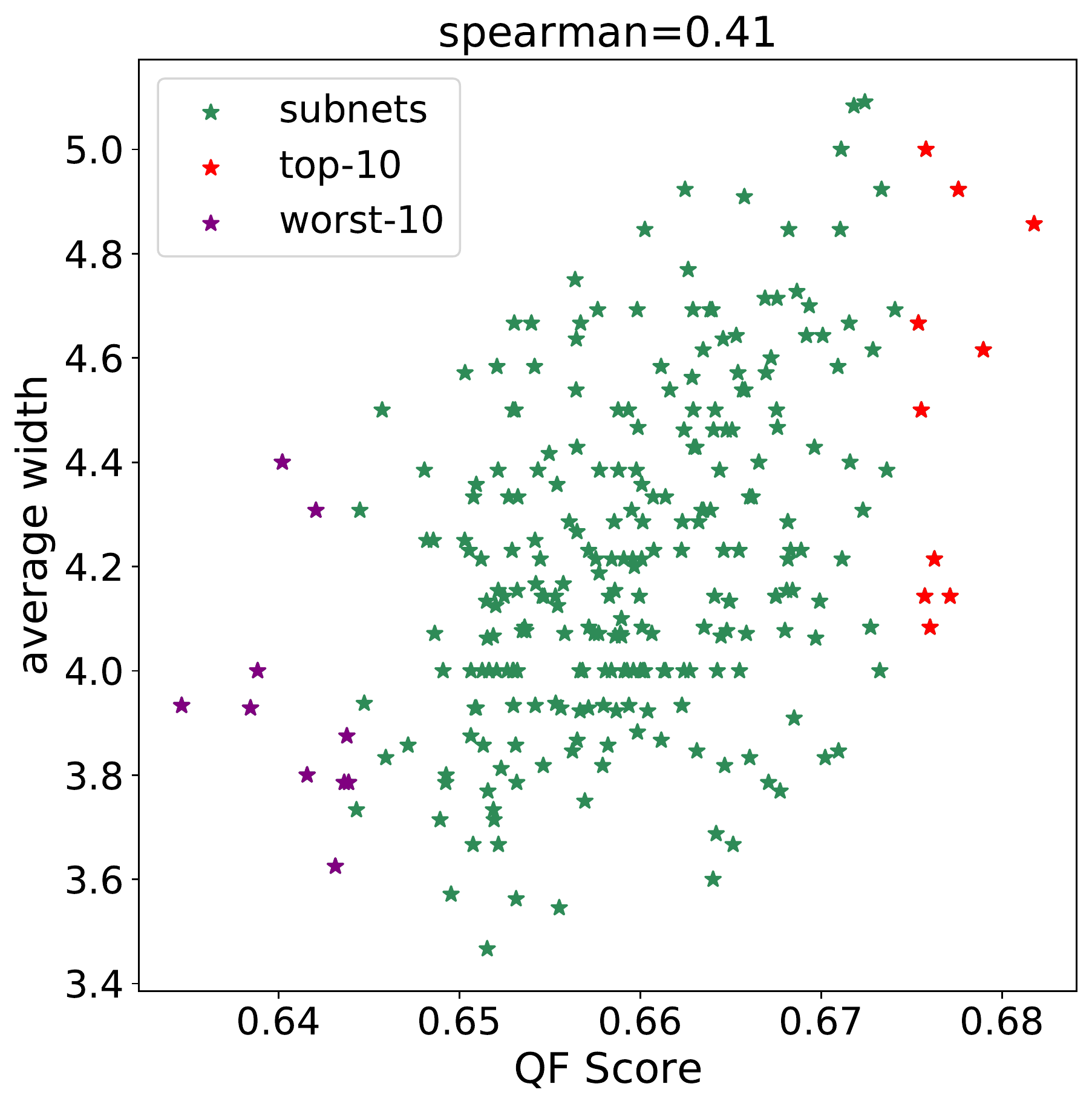}     
} 
\subfigure[Given floating-point accuracy.] { 
\label{fig:fp_acc_flops}     
\includegraphics[width=0.23\linewidth]{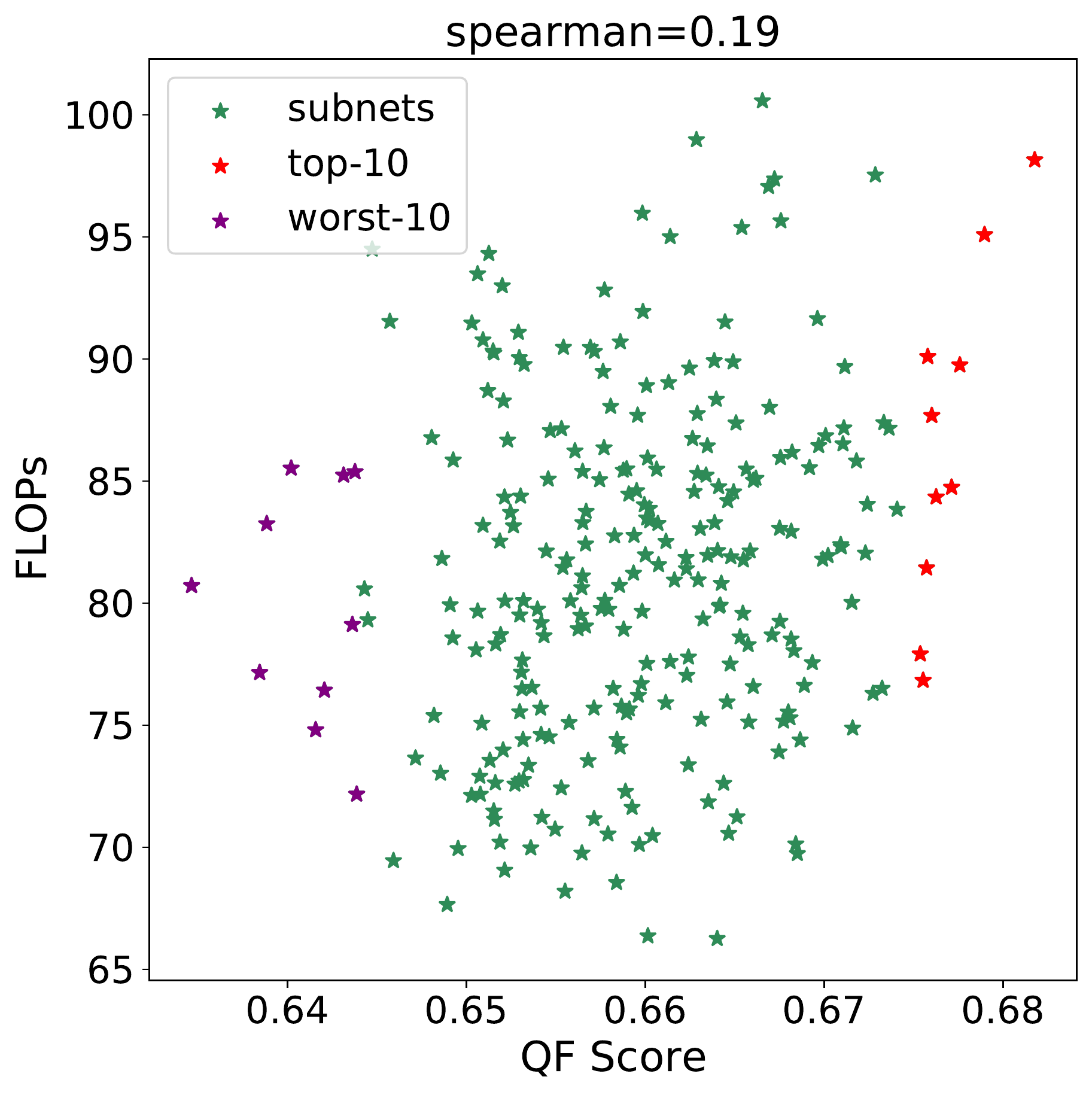}     
}
\caption{Quantization friendly architecture analysis. The quantization score of (b-h) is calculated with the 2bit quantization accuracy. Top-10 and worst-10 are selected with the highest or worst QF scores.}
\label{fig:arch_analysis}   
\end{figure*}

\subsection{Quantization-friendly architecture analysis.} With the large search space and numerous subnets which can be directly deployed, we can analyze the quantization-friendly architectures (QFA) under different bit widths. We sample 20k models from the search space with the flops in the range of $[50M, 300M]$, and we evaluate these architectures from the corresponding supernet to get the accuracy under different bit widths. To evaluate the quantization impact on different architectures, we define a quantization friendliness factor to evaluate QFA under different settings. It is calculated as the ratio $\mathbf{QF_{k}} =\frac{\mathbf {Acc_{k}}}{\mathbf {Acc_{FP}}}$ between the quantization accuracy under $k$ bit $\mathbf {Acc_{k}}$ and floating-point accuracy $\mathbf{Acc_{FP}}$. Spearman correlation coefficient (Spearman) is calculated to measure the correlation between $\mathbf{QF}$ score and elements like flops, resolution, total depth, average width(total width/total depth), and average kernel size(total kernel size/total depth). Only a part of the conclusion is listed as follows.

\paragraph{QFA with different bit-widths.} We analyze the pareto models in the flops and accuracy curve under different bit widths. The distribution of depth, width reveals that the 2-bit models favor shallower depth as shown in Figure~\ref{fig:bit-width_depth}. There is no obvious favored pattern in terms of average width as shown in Figure~\ref{fig:bit-width_width}.

\paragraph{Resolution is important for QFA.} In the visualization of sampled subnets and QF score under 2-bit, $\mathbf{QF}$ score increases greatly from 0.4 to 0.8 as the flops increases as shown in Figure~\ref{fig:flops}. Among all the elements affecting FLOPs, we show that architectures with the large resolution are less sensitive to quantization as shown in Figure~\ref{fig:resolution}, which means that if several architectures with similar flops perform the same in floating-point accuracy, these architectures with large resolution tend to performs better in quantization.

\paragraph{QFA with given computation budget.} We further compare the quantization-friendly pattern with a given computation budget, such as 200M FLOPs (the difference $\leq$ 3\%). We plot 700 architectures in Figure~\ref{fig:computation_budget_depth} and ~\ref{fig:computation_budget_resolution}. It is easy to conclude that the $\mathbf{QF}$ score is negatively correlated to the network depth and positively correlated to the input resolution. The $\mathbf{QF}$ score is almost irrelevant to the average width. Therefore the top-10 models with the highest $\mathbf{QF}$ score have shallower depth than the worst-10 models. It means that with a similar computation budget, we need to design QNN with shallower depth and large resolution.

\paragraph{QFA with the same floating-point accuracy.} Under the same floating-point accuracy $\mathbf{Acc_{FP}}$, the quantization accuracy is decided by $\mathbf{QF}$ score. We plot over 200 architectures with floating-point accuracy $67.8\%$ with (the difference $\leq$ 0.2\%). We find that the average width has a positive correlation with the QF score as shown in Figure~\ref{fig:fp_acc_width}. Besides, the quantization accuracy is less relevant to the FLOPs of models as shown in Figure~\ref{fig:fp_acc_flops}. Therefore, in this case, choosing the wider models results in higher quantization accuracy.

\section{Conclusion}
In this paper, we present Once Quantization-aware Training (OQAT), a framework that deploys the searched quantized models without additional retraining and solves the problem of large accuracy degradation under ultra-low bit widths. With our proposed methods, we can search for the OQATNets model family which far exceeds architectures. Our results reveal the potential of high-performance extremely low-bit neural networks. A comprehensive study reveals the quantization-friendly architectures under different bit widths which might shed a light on further research of high performance extremely low-bit models.


\section{Acknowledgement}
This work was supported by the Australian Research Council Grant DP200103223, and the Australian Medical Research Future Fund MRFAI000085.

\newpage
{\small
\bibliographystyle{ieee_fullname}
\bibliography{egbib}

\begin{thebibliography}{10}\itemsep=-1pt

\bibitem{alizadeh2020gradient}
Milad Alizadeh, Arash Behboodi, Mart van Baalen, Christos Louizos, Tijmen
  Blankevoort, and Max Welling.
\newblock Gradient l1 regularization for quantization robustness.
\newblock {\em arXiv preprint arXiv:2002.07520}, 2020.

\bibitem{bhalgat2020lsq+}
Yash Bhalgat, Jinwon Lee, Markus Nagel, Tijmen Blankevoort, and Nojun Kwak.
\newblock Lsq+: Improving low-bit quantization through learnable offsets and
  better initialization.
\newblock {\em arXiv preprint arXiv:2004.09576}, 2020.

\bibitem{bulat2020bats}
Adrian Bulat, Brais Martinez, and Georgios Tzimiropoulos.
\newblock Bats: Binary architecture search.
\newblock {\em arXiv preprint arXiv:2003.01711}, 2020.

\bibitem{bulat2019xnor}
Adrian Bulat and Georgios Tzimiropoulos.
\newblock Xnor-net++: Improved binary neural networks.
\newblock {\em arXiv preprint arXiv:1909.13863}, 2019.

\bibitem{cai2019once}
Han Cai, Chuang Gan, and Song Han.
\newblock Once for all: Train one network and specialize it for efficient
  deployment.
\newblock {\em arXiv preprint arXiv:1908.09791}, 2019.

\bibitem{cheng2020scalenas}
Hsin-Pai Cheng, Feng Liang, Meng Li, Bowen Cheng, Feng Yan, Hai Li, Vikas
  Chandra, and Yiran Chen.
\newblock Scalenas: One-shot learning of scale-aware representations for visual
  recognition.
\newblock {\em arXiv preprint arXiv:2011.14584}, 2020.

\bibitem{choi2018pact}
Jungwook Choi, Zhuo Wang, Swagath Venkataramani, Pierce I-Jen Chuang,
  Vijayalakshmi Srinivasan, and Kailash Gopalakrishnan.
\newblock Pact: Parameterized clipping activation for quantized neural
  networks.
\newblock {\em arXiv preprint arXiv:1805.06085}, 2018.

\bibitem{deng2009imagenet}
Jia Deng, Wei Dong, Richard Socher, Li-Jia Li, Kai Li, and Li Fei-Fei.
\newblock Imagenet: A large-scale hierarchical image database.
\newblock In {\em 2009 IEEE conference on computer vision and pattern
  recognition}, pages 248--255. Ieee, 2009.

\bibitem{esser2019learned}
Steven~K Esser, Jeffrey~L McKinstry, Deepika Bablani, Rathinakumar Appuswamy,
  and Dharmendra~S Modha.
\newblock Learned step size quantization.
\newblock {\em arXiv preprint arXiv:1902.08153}, 2019.

\bibitem{gong2019differentiable}
Ruihao Gong, Xianglong Liu, Shenghu Jiang, Tianxiang Li, Peng Hu, Jiazhen Lin,
  Fengwei Yu, and Junjie Yan.
\newblock Differentiable soft quantization: Bridging full-precision and low-bit
  neural networks.
\newblock In {\em Proceedings of the IEEE International Conference on Computer
  Vision}, pages 4852--4861, 2019.

\bibitem{guo2019single}
Zichao Guo, Xiangyu Zhang, Haoyuan Mu, Wen Heng, Zechun Liu, Yichen Wei, and
  Jian Sun.
\newblock Single path one-shot neural architecture search with uniform
  sampling.
\newblock {\em arXiv preprint arXiv:1904.00420}, 2019.

\bibitem{he2016deep}
Kaiming He, Xiangyu Zhang, Shaoqing Ren, and Jian Sun.
\newblock Deep residual learning for image recognition.
\newblock In {\em Proceedings of the IEEE conference on computer vision and
  pattern recognition}, pages 770--778, 2016.

\bibitem{howard2019searching}
Andrew Howard, Mark Sandler, Grace Chu, Liang-Chieh Chen, Bo Chen, Mingxing
  Tan, Weijun Wang, Yukun Zhu, Ruoming Pang, Vijay Vasudevan, et~al.
\newblock Searching for mobilenetv3.
\newblock In {\em Proceedings of the IEEE International Conference on Computer
  Vision}, pages 1314--1324, 2019.

\bibitem{hu2018squeeze}
Jie Hu, Li Shen, and Gang Sun.
\newblock Squeeze-and-excitation networks.
\newblock In {\em Proceedings of the IEEE conference on computer vision and
  pattern recognition}, pages 7132--7141, 2018.

\bibitem{hu2021opq}
Peng Hu, Xi Peng, Hongyuan Zhu, Mohamed M~Sabry Aly, and Jie Lin.
\newblock Opq: Compressing deep neural networks with one-shot
  pruning-quantization.
\newblock In {\em Proceedings of the AAAI Conference on Artificial
  Intelligence}, pages 7780--7788, 2021.

\bibitem{jin2019adabits}
Qing Jin, Linjie Yang, and Zhenyu Liao.
\newblock Adabits: Neural network quantization with adaptive bit-widths.
\newblock {\em arXiv preprint arXiv:1912.09666}, 2019.

\bibitem{kim2019qkd}
Jangho Kim, Yash Bhalgat, Jinwon Lee, Chirag Patel, and Nojun Kwak.
\newblock Qkd: Quantization-aware knowledge distillation.
\newblock {\em arXiv preprint arXiv:1911.12491}, 2019.

\bibitem{li2019lfs}
Chuming Li, Xin Yuan, Chen Lin, Minghao Guo, Wei Wu, Junjie Yan, and Wanli
  Ouyang.
\newblock Am-lfs: Automl for loss function search.
\newblock In {\em Proceedings of the IEEE/CVF International Conference on
  Computer Vision}, pages 8410--8419, 2019.

\bibitem{li2020improving}
Xiang Li, Chen Lin, Chuming Li, Ming Sun, Wei Wu, Junjie Yan, and Wanli Ouyang.
\newblock Improving one-shot nas by suppressing the posterior fading.
\newblock In {\em Proceedings of the IEEE/CVF Conference on Computer Vision and
  Pattern Recognition}, pages 13836--13845, 2020.

\bibitem{li2019additive}
Yuhang Li, Xin Dong, and Wei Wang.
\newblock Additive powers-of-two quantization: An efficient non-uniform
  discretization for neural networks.
\newblock {\em arXiv preprint arXiv:1909.13144}, 2019.

\bibitem{li2021brecq}
Yuhang Li, Ruihao Gong, Xu Tan, Yang Yang, Peng Hu, Qi Zhang, Fengwei Yu, Wei
  Wang, and Shi Gu.
\newblock Brecq: Pushing the limit of post-training quantization by block
  reconstruction.
\newblock {\em arXiv preprint arXiv:2102.05426}, 2021.

\bibitem{li2021mqbench}
Yuhang Li, Mingzhu Shen, Jian Ma, Yan Ren, Mingxin Zhao, Qi Zhang, Ruihao Gong,
  Fengwei Yu, and Junjie Yan.
\newblock {MQB}ench: Towards reproducible and deployable model quantization
  benchmark.
\newblock In {\em Thirty-fifth Conference on Neural Information Processing
  Systems Datasets and Benchmarks Track (Round 1)}, 2021.

\bibitem{li2020efficient}
Yuhang Li, Wei Wang, Haoli Bai, Ruihao Gong, Xin Dong, and Fengwei Yu.
\newblock Efficient bitwidth search for practical mixed precision neural
  network.
\newblock {\em arXiv preprint arXiv:2003.07577}, 2020.

\bibitem{liang2019computation}
Feng Liang, Chen Lin, Ronghao Guo, Ming Sun, Wei Wu, Junjie Yan, and Wanli
  Ouyang.
\newblock Computation reallocation for object detection.
\newblock {\em arXiv preprint arXiv:1912.11234}, 2019.

\bibitem{liu2018darts}
Hanxiao Liu, Karen Simonyan, and Yiming Yang.
\newblock Darts: Differentiable architecture search.
\newblock {\em arXiv preprint arXiv:1806.09055}, 2018.

\bibitem{liu2020reactnet}
Zechun Liu, Zhiqiang Shen, Marios Savvides, and Kwang-Ting Cheng.
\newblock Reactnet: Towards precise binary neural network with generalized
  activation functions.
\newblock In {\em European Conference on Computer Vision (ECCV)}, 2020.

\bibitem{mishra2017wrpn}
Asit Mishra, Eriko Nurvitadhi, Jeffrey~J Cook, and Debbie Marr.
\newblock Wrpn: wide reduced-precision networks.
\newblock {\em arXiv preprint arXiv:1709.01134}, 2017.

\bibitem{nagel2020up}
Markus Nagel, Rana~Ali Amjad, Mart Van~Baalen, Christos Louizos, and Tijmen
  Blankevoort.
\newblock Up or down? adaptive rounding for post-training quantization.
\newblock In {\em International Conference on Machine Learning}, pages
  7197--7206. PMLR, 2020.

\bibitem{phan2020binarizing}
Hai Phan, Zechun Liu, Dang Huynh, Marios Savvides, Kwang-Ting Cheng, and
  Zhiqiang Shen.
\newblock Binarizing mobilenet via evolution-based searching.
\newblock {\em arXiv preprint arXiv:2005.06305}, 2020.

\bibitem{sandler2018mobilenetv2}
Mark Sandler, Andrew Howard, Menglong Zhu, Andrey Zhmoginov, and Liang-Chieh
  Chen.
\newblock Mobilenetv2: Inverted residuals and linear bottlenecks.
\newblock In {\em Proceedings of the IEEE conference on computer vision and
  pattern recognition}, pages 4510--4520, 2018.

\bibitem{shen2019searching}
Mingzhu Shen, Kai Han, Chunjing Xu, and Yunhe Wang.
\newblock Searching for accurate binary neural architectures.
\newblock In {\em Proceedings of the IEEE International Conference on Computer
  Vision Workshops}, pages 0--0, 2019.

\bibitem{shkolnik2020robust}
Moran Shkolnik, Brian Chmiel, Ron Banner, Gil Shomron, Yuri Nahshan, Alex
  Bronstein, and Uri Weiser.
\newblock Robust quantization: One model to rule them all.
\newblock {\em arXiv preprint arXiv:2002.07686}, 2020.

\bibitem{tan2019efficientnet}
Mingxing Tan and Quoc~V Le.
\newblock Efficientnet: Rethinking model scaling for convolutional neural
  networks.
\newblock {\em arXiv preprint arXiv:1905.11946}, 2019.

\bibitem{wang2019haq}
Kuan Wang, Zhijian Liu, Yujun Lin, Ji Lin, and Song Han.
\newblock Haq: Hardware-aware automated quantization with mixed precision.
\newblock In {\em Proceedings of the IEEE conference on computer vision and
  pattern recognition}, pages 8612--8620, 2019.

\bibitem{wang2020apq}
Tianzhe Wang, Kuan Wang, Han Cai, Ji Lin, Zhijian Liu, Hanrui Wang, Yujun Lin,
  and Song Han.
\newblock Apq: Joint search for network architecture, pruning and quantization
  policy.
\newblock In {\em Proceedings of the IEEE/CVF Conference on Computer Vision and
  Pattern Recognition}, pages 2078--2087, 2020.

\bibitem{wu2018mixed}
Bichen Wu, Yanghan Wang, Peizhao Zhang, Yuandong Tian, Peter Vajda, and Kurt
  Keutzer.
\newblock Mixed precision quantization of convnets via differentiable neural
  architecture search.
\newblock {\em arXiv preprint arXiv:1812.00090}, 2018.

\bibitem{yu2019universally}
Jiahui Yu and Thomas~S Huang.
\newblock Universally slimmable networks and improved training techniques.
\newblock In {\em Proceedings of the IEEE International Conference on Computer
  Vision}, pages 1803--1811, 2019.

\bibitem{yu2020bignas}
Jiahui Yu, Pengchong Jin, Hanxiao Liu, Gabriel Bender, Pieter-Jan Kindermans,
  Mingxing Tan, Thomas Huang, Xiaodan Song, Ruoming Pang, and Quoc Le.
\newblock Bignas: Scaling up neural architecture search with big single-stage
  models.
\newblock {\em arXiv preprint arXiv:2003.11142}, 2020.

\bibitem{yu2018slimmable}
Jiahui Yu, Linjie Yang, Ning Xu, Jianchao Yang, and Thomas Huang.
\newblock Slimmable neural networks.
\newblock {\em arXiv preprint arXiv:1812.08928}, 2018.

\bibitem{zhou2020econas}
Dongzhan Zhou, Xinchi Zhou, Wenwei Zhang, Chen~Change Loy, Shuai Yi, Xuesen
  Zhang, and Wanli Ouyang.
\newblock Econas: Finding proxies for economical neural architecture search.
\newblock In {\em Proceedings of the IEEE/CVF Conference on Computer Vision and
  Pattern Recognition}, pages 11396--11404, 2020.

\bibitem{zhou2016dorefa}
Shuchang Zhou, Yuxin Wu, Zekun Ni, Xinyu Zhou, He Wen, and Yuheng Zou.
\newblock Dorefa-net: Training low bitwidth convolutional neural networks with
  low bitwidth gradients.
\newblock {\em arXiv preprint arXiv:1606.06160}, 2016.

\end{thebibliography}
}

\end{document}